\documentclass{article}

\usepackage{iclr2027_conference,times}
\usepackage[utf8]{inputenc}
\usepackage[T1]{fontenc}
\usepackage{url}
\usepackage{booktabs}
\usepackage{amsmath}
\usepackage{amsfonts}
\usepackage{graphicx}
\usepackage{float}
\usepackage{microtype}
\usepackage{bbm}
\usepackage{tabularx}
\usepackage{array}

\usepackage{amssymb}

\usepackage[table]{xcolor}

\usepackage{fontawesome5}
\definecolor{linkblue}{RGB}{125,142,185}

\usepackage{hyperref}

\definecolor{excessLow}{RGB}{253,239,237}
\definecolor{excessMidLow}{RGB}{251,229,225}
\definecolor{excessMidHigh}{RGB}{249,219,214}
\definecolor{excessHigh}{RGB}{246,207,201}
\definecolor{nullCell}{RGB}{243,243,243}
\definecolor{sigRed}{RGB}{180,45,38}

\newcommand{\cmark}{\ensuremath{\checkmark}}
\newcommand{\tmark}{\ensuremath{\triangle}}
\newcommand{\xmark}{\ensuremath{\times}}
\newcommand{\pp}{\ensuremath{\,\mathrm{pp}}}

\newcolumntype{Z}{>{\centering\arraybackslash}X}

\title{Are Near-Tied LLM Rankings Robust to Family-DIF-Guided Benchmark Recomposition?}

\author{%
\mdseries
\begin{tabular}{@{}l@{\hspace{4em}}c@{}}
\begin{tabular}{@{}l@{}}
\textbf{Qiaoyuan Zheng, Yiqu Yang} \\
Department of Computer Science, ETH Zurich, Switzerland \\
\texttt{\{zqiaoyuan, yangyiq\}@ethz.ch}
\end{tabular}
&
\href{https://github.com/qiaoyuan667/near-tie-robustness}{%
  \faGithub\enspace\textcolor{linkblue}{GitHub}%
}
\end{tabular}%
}

\date{}
\iclrfinalcopy
\begin{document}

\maketitle
\lhead{Preprint}

\begin{abstract}
Small leaderboard gaps are often interpreted as evidence that one language model is better than another, but their sign may depend on which benchmark items are included. We test this using item-level responses from five benchmarks and a family-label-free spectral approximation to multidimensional item-response theory (MIRT). In owner-disjoint folds, one owner half identifies items with low residual differential item functioning across model families (low-DIF). These items are used to score models in the other half with frozen weights that preserve the benchmark's composition across metadata-defined item groups and easiness strata. Equally short matched-random subtests provide a baseline for variation due to item subsampling. Full-benchmark and low-DIF rankings remain strongly correlated ($\tau_b=.900$--$.948$). Yet in four of five benchmarks, 30.9--47.1\% of cross-family pairs initially within one percentage point reverse order, exceeding their matched-random medians by 16.9--28.6 percentage points (all $p=.001$). The fifth benchmark shows no reliable excess ($-0.9$ points, $p=.689$). The pattern survives all pre-specified population perturbations, and residual item--family signatures replicate across owner halves; however, no family shows a consistent advantage across benchmarks. Thus, globally stable rankings can still leave individual near-tie orderings sensitive to benchmark composition, and sub-one-point leaderboard gaps should be accompanied by evidence that the implied ordering is composition-robust.
\end{abstract}

\section{Introduction}

\begin{figure}[b]
  \centering
  \includegraphics[width=\textwidth]{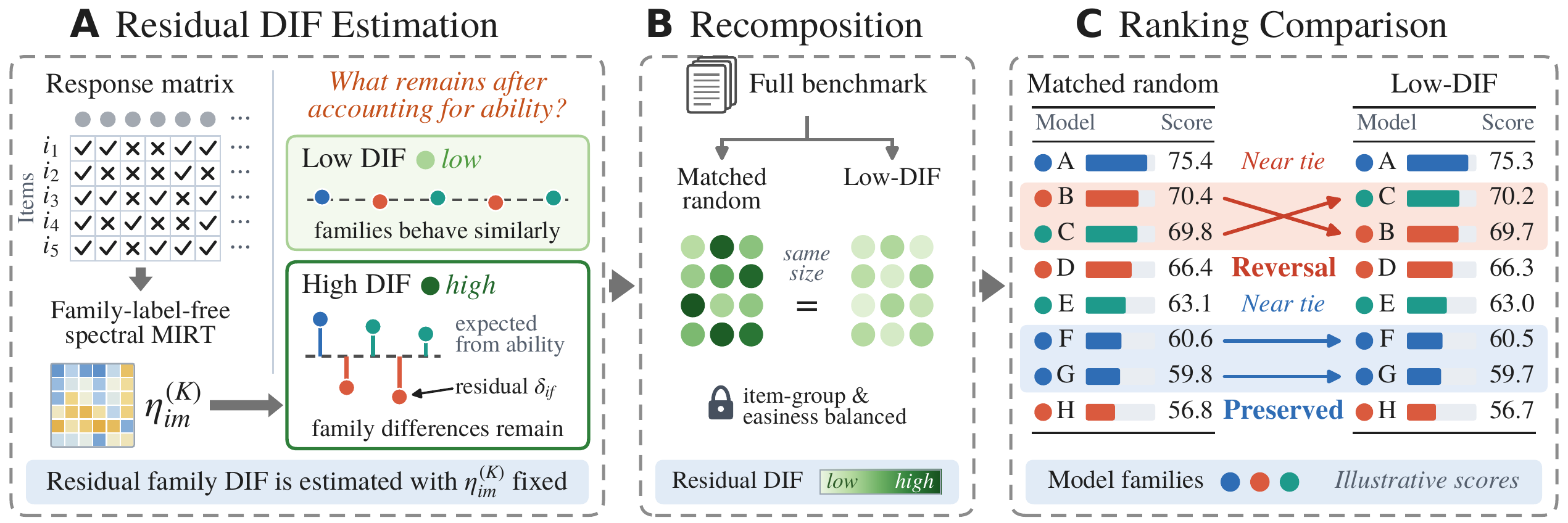}
  \caption{\textbf{(A)} Residual DIF flags items that favor some model families at equal ability.
\textbf{(B)} Low-DIF items are recomposed into a subset matched to a random one.
\textbf{(C)} Near-tie rankings can reverse.}
  \label{fig:overview}
\end{figure}

Static, closed-form benchmark leaderboards collapse thousands of item-level
responses into a single score, often separating models by less than one
percentage point. Such gaps do not establish that the implied ordering is robust to benchmark
item composition. Rankings can
change under prompt weighting or evaluation perturbations
\citep{siska2024distribution,alzahrani2024targets}, while conventional
uncertainty analyses leave some close comparisons unresolved
\citep{kotawala2026resolution}. We ask whether a near-tied ordering is robust
to which benchmark items are included.

We study this question through family-dependent item functioning (family-DIF) as shown in Figure~\ref{fig:overview},
adapting differential item functioning from psychometrics
\citep{holland1993differential}. After a family-blind multidimensional
adjustment, residual item--family effects measure systematic differences across
observational model families. In owner-disjoint folds, one owner half selects low-DIF anchors balanced across metadata-defined item groups---such as MMLU subjects and BBH tasks---and easiness strata; the resulting frozen weights score the other half, after which the halves are swapped. Equally short matched-random subtests separate targeted composition sensitivity from generic shorter-subtest variation.

Despite stable aggregate rankings, four of five benchmarks show excess
near-tie reversals over matched-random controls. A post-hoc pair-type
diagnostic separates a generic within-family recomposition component from an
additional cross-family component; the latter is significant in all four
primary-positive benchmarks.
No family shows a consistent advantage across benchmarks, suggesting that the observed reversals cannot be attributed to a single model family systematically benefiting from the recomposition procedure.
\section{Related work}

\paragraph{Psychometrics for model evaluation.}
IRT has been used to construct evaluation scales for NLP systems
\citep{lalor2016building}, assess leaderboard and test-set
discriminability and identify informative examples
\citep{rodriguez2021evaluation,vania2021comparing}, detect invalid or
mislabeled benchmark items
\citep{truong2025fantastic,land2026auditing}, and reduce evaluation
cost
\citep{polo2024tinybenchmarks,hofmann2025fluid,zhou2026lost}. Recent methods learn
text-conditioned IRT representations for model routing and benchmark prediction
\citep{chen2025irtnet}, or use fixed-parameter MIRT anchors to preserve
comparability as benchmark suites evolve \citep{habba2026growing}. In contrast,
our spectral representation controls for multidimensional response structure,
and our low-DIF anchors diagnose composition sensitivity rather than predict
performance, link test forms, or detect mislabeled items.

Prior DIF work examines human--chatbot differences
\citep{zeinfeld2026assessment} and scoring procedures that downweight DIF items
\citep{halpin2024dtf}. A course report closest to our setting finds that DIF-selected BBH pools alter
held-out Llama--Qwen group gaps \citep{liu2026bbhdif}. We extend this
baseline to owner-disjoint rankings across five benchmarks, individual near-tied
pairs, and random controls matched on item group and easiness.

\paragraph{Ranking uncertainty and benchmark robustness.}
Model rankings are sensitive to item weighting, prompt composition, and
evaluation protocols
\citep{mishra2021robust,siska2024distribution,kim2026benchhub,
alzahrani2024targets}. Related work augments accuracy with model-output uncertainty
\citep{ye2024uncertainty}, quantifies variance across training and
evaluation choices \citep{madaan2024variance}, shows that clustered
benchmark structure can widen rank uncertainty
\citep{neuhof2026ranking}, and tests whether adjacent model
comparisons have adequate paired statistical resolution
\citep{kotawala2026resolution}. \citet{heineman2025signal} analyze ranking fidelity
under training and checkpoint noise, while \citet{qian2026benchmark2} use
discriminability and within-family inversions for subset selection. Our distinguishing target is excess reversal of individual cross-family near ties under owner-disjoint DIF-guided recomposition, measured against equally short, composition-matched random subtests (Appendix~\ref{app:related-work-comparison}).

\paragraph{Measurement validity and interpretation.}
Benchmark validity depends on the inferences supported by the measurement
protocol \citep{bean2025construct}. Multidimensional capability structure can provide an alternative explanation
for apparent DIF \citep{zhou2026generalscales}, while conventional IRT
estimators can be unreliable in AI benchmark regimes with few models or
non-normal ability distributions \citep{jiang2026trust}. Diagnostic conclusions
can also be fragile \citep{aribandi2021reliable}, and textual correlates of DIF
may reflect intended subdomains rather than construct-irrelevant content
\citep{maeda2025finding}. We therefore
separate ranking consequences, owner-disjoint stability, and blinded semantic
interpretation. Our residual effects remain conditional on a family-blind,
prediction-selected spectral response representation and do not establish that
every capability difference has been removed.

\section{Method}
\label{sec:method}

We test one form of composition robustness: whether near-tied rankings
survive family-DIF-guided benchmark recomposition. One owner half
identifies low-DIF items and constructs weights that preserve item-group
and easiness composition; the frozen weights then score models in the
opposite half. Matched-random subtests provide a baseline for variation
due to item subsampling. Implementation and dimension-selection settings
are summarized in Table~\ref{tab:v3-hyperparameters}.

\subsection{Data and Models}

For each benchmark, we have a binary response matrix
$Y\in\{0,1\}^{I\times M}$, where $Y_{im}=1$ if model $m$
answers item $i$ correctly. We audit MMLU-Pro
\citep{wang2024mmlupro}, BIG-Bench Hard
\citep{suzgun2023bbh}, MMLU \citep{hendrycks2021mmlu},
HellaSwag \citep{zellers2019hellaswag}, and WinoGrande
\citep{sakaguchi2021winogrande}, using the item-level response
matrices released with RouterEval \citep{huang2025routereval}.

Models are assigned to Qwen, Llama, Gemma, Mistral/Mixtral, or Phi
via a frozen, unique model-name match. We cap each owner--family
pair at five checkpoints and assign entire owners to two approximately
balanced, disjoint halves. For ranking comparisons, we retain one
checkpoint per owner--family pair: the checkpoint whose full-benchmark
accuracy is closest to that pair's median. Across benchmarks, the complete-response matrices contain
1,267--14,042 items and 1,823--5,000 models, while the ranking
analyses use 204--673 owner--family representatives per benchmark.
Appendix~\ref{app:population-construction} details model-name matching,
owner extraction, capping, and fold construction.

\subsection{Family DIF Estimation}

\paragraph{Capability adjustment.}
We use a family-label-free spectral working model to approximate the
multidimensional item--model term in compensatory MIRT
\citep{reckase2009mirt}. This provides predictive nuisance adjustment,
rather than latent-trait interpretation or traditional marginal
maximum-likelihood estimation. It permits tractable repeated
owner- and item-held-out fitting and anchor purification without a
parametric population distribution for model abilities. The four-benchmark excess-reversal pattern also holds under direct penalized joint logistic MIRT (Appendix~\ref{app:direct-mirt}).
Let $\eta^{(K)}_{im}$ denote the rank-$K$ reconstruction approximating
$\mathbf a_i^\top\boldsymbol\theta_m$. For item $i$ and model $m$
from observational family $f(m)$, the linear predictor is
\begin{equation}
\xi_{im}
=
\eta^{(K)}_{im}
+b_i
+\delta_{i,f(m)},
\label{eq:dif-linear}
\end{equation}
with response probability
\begin{equation}
\Pr(Y_{im}=1)
=
\sigma(\xi_{im}),
\qquad
\sigma(x)=\frac{1}{1+e^{-x}},
\label{eq:mirt-dif}
\end{equation}
where $b_i$ is item easiness and $\delta_{i,f(m)}$ is the residual
item-by-family effect.
Within each outer owner half, nested owner-held-out and item-held-out
Bernoulli log loss and the one-standard-error rule
\citep{hastie2009elements} select $K$ from
$\{1,2,4,8,16,32,64,128,256\}$, omitting values larger than the
smaller dimension of the corresponding inner-training fitting matrix.
All downstream analyses use this candidate grid. Family labels are
not used to learn the response directions or select $K$, and the
opposite owner half does not select the directions, dimension, or
anchor items. Appendix~\ref{app:mirt-dimension} details the spectral
construction, held-out coordinate estimation, and dimension selection.

\paragraph{Residual family DIF.}
With $\eta^{(K)}_{im}$ fixed, we estimate item easiness and residual
family effects by
\begin{equation}
(\widehat b,\widehat\delta)
=
\arg\min_{b,\delta}
\left\{
\sum_{i,m}
\left[
\log(1+e^{\xi_{im}})
-Y_{im}\xi_{im}
\right]
+
\frac{\lambda}{2}\sum_{i,f}\delta_{if}^{2}
\right\},
\qquad \lambda=1.
\label{eq:dif-objective}
\end{equation}
For identifiability, family effects are centered within each item:
\begin{equation}
\sum_f\delta_{if}=0
\qquad
\text{for every item } i.
\label{eq:dif-centering}
\end{equation}
Thus, $\delta_{if}$ is family $f$'s conditional deviation from the
item's across-family mean. We summarize residual family dependence by
\begin{equation}
D_i
=
\max_f\widehat\delta_{if}
-
\min_f\widehat\delta_{if}.
\label{eq:dif-magnitude}
\end{equation}
A large $D_i$ indicates separation between observational families
after conditioning on the selected spectral approximation, not item
unfairness or a causal family mechanism. Optimization and convergence
details appear in Appendix~\ref{app:dif-optimization}.

\subsection{Low-DIF Anchor Construction}

\paragraph{Owner-disjoint scoring.}
Here, an ``anchor'' is a diagnostic scoring subset, not a scale-linking
anchor used to preserve an IRT scale across evolving test forms
\citep{habba2026growing}.
We use two-fold, owner-disjoint cross-fitting in the sample-splitting
sense \citep{chernozhukov2018double} to avoid selecting items using
responses from the evaluated model's owner. In each direction, one
owner half selects the spectral MIRT dimension, learns the response
directions, estimates residual family DIF, and constructs anchor
weights. These weights are frozen and applied to representative models
from the opposite half. Swapping the halves ensures that every
evaluated model is scored using anchors selected without its owner's
responses.

\paragraph{Purification and item selection.}
Starting with all items, we run three fixed purification rounds,
following the logic of iterative DIF purification
\citep{candell1988iterative}. Each round learns spectral MIRT response
directions from the current anchors, projects all items onto those
directions, and re-estimates residual family DIF on all items.
Within each metadata-defined item group, items are ordered by fitted
residual easiness $b_i$ and divided into up to four nonempty,
near-equal strata. Each resulting subset is a \emph{blueprint cell}.
Within each cell, we retain the fraction $\alpha$ of items with the
smallest $D_i$: $\alpha=.50$ in the primary analysis, with
$\alpha=.30$ and $\alpha=.70$ as sensitivity analyses.
The third round's selection and weights are frozen.
Appendix~\ref{app:purification} specifies deterministic within-cell
selection, the full three-round procedure, and final refitting.

\paragraph{Blueprint-preserving weights.}
For a cell $c$ containing $n_c$ items and $k_c$ selected anchors,
we assign weights
\begin{equation}
w_i
=
\begin{cases}
n_c/k_c,
& i \text{ is a selected anchor in cell } c,\\
0,
& \text{otherwise}.
\end{cases}
\label{eq:anchor-weight}
\end{equation}
Consequently,
\begin{equation}
\sum_{i\in c}w_i=n_c,
\label{eq:cell-mass}
\end{equation}
so each cell retains its original total weight. Low-DIF selection
therefore changes residual-DIF composition while preserving the joint
distribution of item groups and coarse easiness strata under the
scoring weights. For a held-out model $m$, the full-benchmark and
weighted anchor scores are
\begin{equation}
A_m
=
\frac{1}{I}\sum_iY_{im},
\qquad
\widetilde A_m
=
\frac{\sum_iw_iY_{im}}{\sum_iw_i}.
\label{eq:scores}
\end{equation}

\subsection{Ranking estimand and matched control}
\label{sec:controls}

\paragraph{Near-tie reversals.}
The primary comparison set comprises different-owner,
different-family model pairs satisfying $|A_m-A_n|\leq .01$.
A strict rank reversal occurs when
\begin{equation}
(A_m-A_n)(\widetilde A_m-\widetilde A_n)<0.
\label{eq:reversal}
\end{equation}
Pairs tied under either scoring rule are excluded from the
corresponding denominator. We pool eligible-pair and reversal counts
across the two held-out owner folds before computing the reversal
rate. Secondary summaries include mean fold-specific Kendall
$\tau_b$ \citep{kendall1945ties}, the all-cross-family reversal rate,
and cumulative results at score-gap thresholds
$\{0.25,0.5,1,2,3,5\}$ percentage points.

\paragraph{Matched-random controls.}
To assess variation from item subsampling, we compare the low-DIF
anchor with $R=1{,}000$ blueprint-matched random subtests.
Within each outer fold, every replicate samples without replacement
the same number of items from each blueprint cell as the anchor and
applies the same cell-restoring weights. The controls match test
length and the joint composition of item groups and coarse easiness
strata without selecting items by $D_i$. Replicate reversal counts
are pooled across folds in the same way as the low-DIF counts
(Appendix~\ref{app:random-control}).
Let $T_{\mathrm{DIF}}$ be the observed pooled reversal rate and $T_r$
the rate for random replicate $r$. We report excess reversals as
\begin{equation}
\Delta_{\mathrm{pp}}
=
100\left(
T_{\mathrm{DIF}}
-
\operatorname{median}_{1\leq r\leq R}T_r
\right),
\label{eq:excess-reversal}
\end{equation}
with the one-sided empirical randomization $p$-value
\citep{phipson2010permutation}
\begin{equation}
p
=
\frac{
1+\sum_{r=1}^{R}\mathbf{1}\{T_r\geq T_{\mathrm{DIF}}\}
}{
R+1
}.
\label{eq:random-control}
\end{equation}
Thus, $\Delta_{\mathrm{pp}}>0$ means that low-DIF scoring reverses
near-tied pairs more often than equally short, blueprint-matched
random subtests.

\paragraph{Composition sensitivity.}
Separately, 2,000 item-group block-bootstrap replicates
\citep{efron1993bootstrap} assess sensitivity to benchmark
recomposition. Single-item-group benchmarks use 100 deterministic
contiguous item clusters. Appendix~\ref{app:composition-bootstrap} reports the 2.5th--97.5th percentile ranges of
the recomposed near-tie reversal rate and mean fold-specific
Kendall correlation. The near-tie pair set is frozen from the observed
full-benchmark scores throughout all random-control and resampling
analyses.

\section{Validation and Diagnostics}
\label{sec:validation}

We examine whether the ranking effect can be explained by discrimination or
reliability differences, whether it is stronger across than within model
families, and whether it persists across changes in the model population. We
also test whether residual item--family effects replicate across owner halves
and are associated with broad content categories.

\subsection{Discrimination and Reliability Controls}
\label{sec:discrimination-control}

The blueprint-matched random subtests used in the primary analysis
(Section~\ref{sec:controls}) preserve test length, item-group composition, and
coarse easiness composition, but do not explicitly match item discrimination
or test reliability. We therefore conduct an additional diagnostic analysis
that matches controls on audit-half predicted-logit spread within each blueprint cell and reports balance in corrected item--total
correlation, standard error of measurement, and Cronbach's $\alpha$.
The construction uses only the audit half; control weights are frozen before
scoring the opposite target half. Full definitions appear in
Appendix~\ref{app:discrimination-control-method}.

\subsection{Within-family specificity diagnostic}
\label{sec:within-family-specificity}

The primary estimand concerns different-owner, different-family near ties.
To distinguish a generic sensitivity to low-DIF recomposition from an
additional family-specific component, we conduct a post-hoc pair-type
diagnostic. We replace the different-family restriction with a same-family
restriction while retaining different owners, the frozen one-percentage-point
full-score band, strict tie exclusions, the frozen 50\% low-DIF weights, and
the same 1,000 blueprint-matched random subtests.

For every random replicate, we calculate both cross-family and within-family
reversal rates. The specificity statistic compares the observed
cross-minus-within reversal gap with the paired distribution of
cross-minus-within gaps from the same matched-random subtests. Full definitions
and results appear in Appendix~\ref{app:within-family-diagnostic}.

\subsection{Model Population Robustness}
\label{sec:robustness}

We test sensitivity to owner concentration, checkpoint selection,
lineage definition, and family composition. Relative to the
owner-cap-5 baseline, the nine perturbations use owner caps of 1 and
3, a score-blind lexicographic representative, a conservative
name-based lineage exclusion, or omit Gemma, Llama, Mistral/Mixtral,
Phi, or Qwen in turn.

Each variant preserves the owner-disjoint cross-fit and refits the
spectral response directions, residual DIF, anchors, and matched
controls; only the benchmark- and fold-specific dimension $K$ from
the primary analysis is frozen. The frozen criteria require every perturbation to retain positive excess
reversals in at least three of five benchmarks and positive excess reversals
with matched-random $p\leq .05$ in at least three. New variants use 200 random controls; the baseline
uses 1,000. Full specifications appear in
Appendix~\ref{app:population-robustness}.

\subsection{Item--Family Effect Replication}
\label{sec:item-stability}

We refit residual item--family effects separately in the two owner halves
and construct shared blueprint cells from their averaged item
intercepts. After within-cell adjustment, we report three cross-half
replication statistics: the median family-wise Spearman correlation, top-20\%
high-DIF overlap, and advantaged-family agreement among overlapping items.
Under independence, the correlation is centered at zero and the expected
top-20\% overlap is 20\%; agreement has no fixed baseline because family
frequencies differ. We therefore assess all three statistics using 500
one-sided within-cell permutations and report their effect sizes and
$p$-values.

For MMLU-Pro, BBH, and MMLU, secondary analyses test whether
discovery-selected extreme item-group effects reproduce in validation and whether
exact item-group contributions to family score shifts retain their signs across
folds. Full definitions appear in
Appendix~\ref{app:item-stability-source}.

\subsection{Blinded Content Audit}
\label{sec:content-audit}

To interpret the reproducible item--family effects, we sampled 50 stable
high-DIF items per benchmark and paired each with a control from the same blueprint cell, yielding 250 matched pairs. High-DIF items were in the
within-cell top 20\% in both owner halves; controls were in neither top-20\%
set.

Two independently prompted language-model annotators, blinded to benchmark,
group, DIF, and family information, labeled ten binary and three ordinal
content axes. Their labels were analyzed separately.

A binary axis was considered confirmatory only if both annotators showed the
same pooled direction, BH $q\leq .05$, an absolute paired difference of at
least eight percentage points, and the same direction in at least three
benchmarks. Family-direction analyses were post-confirmatory and exploratory.
Full details appear in Appendix~\ref{app:content-audit}.

\section{Results}
\subsection{Sensitivity to Benchmark Composition}

\paragraph{Stable overall rankings and near-tie reversals.}

Table~\ref{tab:main} reports results using the family-label-free spectral MIRT
approximation. Full-benchmark and low-DIF anchor rankings remain strongly
correlated ($\tau_b=.900$--$.948$), with only 2.5--4.2\% of all cross-family
comparisons reversing. Among pairs initially within one percentage point,
however, four benchmarks show reversal rates of 30.9--47.1\%, exceeding
matched-random medians by 16.9--28.6 percentage points
(all matched-random $p=.001$).

For WinoGrande, low-DIF and matched-random reversal rates are similar
(33.8\% vs.\ 34.7\%; excess $-0.9$ points, $p=.689$), providing no evidence
of excess beyond its high generic subtest-variation baseline. In the other
four benchmarks, excess sensitivity is concentrated among near-tied
comparisons even as aggregate rankings remain stable. Near-tie reversals
remain substantial across alternative anchor fractions
(Appendix~\ref{app:anchor-fraction-sensitivity}).

\paragraph{Discrimination and reliability control.}

Discrimination-matched controls closely reproduce the measurement properties
of the low-DIF subtests. Nevertheless, the four primary-positive benchmarks retain
excess reversals of 14.2--24.0 percentage points (all $p=.001$), while
WinoGrande remains null ($-0.8$ points, $p=.683$;
Table~\ref{tab:discrimination-matched-control}). The primary result is
therefore not explained by lower aggregate discrimination or reliability
under these diagnostics.

\begin{table}[t]
\centering
\caption{
Primary near-tie ranking results using the 50\% residual low-DIF
anchors. $K$ gives the selected spectral MIRT dimensions in the two
audit folds. $\tau_b$ is the mean fold-specific Kendall correlation
between the full-benchmark and low-DIF rankings. The Near-tie pairs column reports the pooled number of model pairs
from different owners and families whose full-benchmark scores differ
by at most one percentage point, excluding ties under low-DIF scoring.
Low-DIF is the pooled strict reversal rate for these pairs under anchor
scoring; Random is the median reversal rate across 1,000
blueprint-matched random subtests. Excess is Low-DIF minus Random in
percentage points, and $p_{\mathrm{rand}}$ is the corresponding
one-sided empirical matched-random $p$-value.
}
\label{tab:main}

\small
\setlength{\tabcolsep}{4.5pt}
\renewcommand{\arraystretch}{1.08}

\begin{tabular}{lccccccc}
\toprule
Benchmark
& $K$
& $\tau_b$
& Near-tie pairs
& Low-DIF (\%)
& Random (\%)
& Excess (pp)
& $p_{\mathrm{rand}}$ \\
\midrule

MMLU-Pro
& 64/32
& .924
& 308
& 47.1
& 18.5
& \cellcolor{excessHigh}+28.6
& \textcolor{sigRed}{.001}
\\

BBH
& 128/128
& .915
& 2,226
& 42.1
& 25.2
& \cellcolor{excessLow}+16.9
& \textcolor{sigRed}{.001}
\\

MMLU
& 128/128
& .930
& 3,235
& 40.7
& 16.3
& \cellcolor{excessMidHigh}+24.4
& \textcolor{sigRed}{.001}
\\

HellaSwag
& 16/64
& .948
& 4,533
& 30.9
& 11.4
& \cellcolor{excessMidLow}+19.5
& \textcolor{sigRed}{.001}
\\

WinoGrande
& 8/8
& .900
& 5,574
& 33.8
& 34.7
& \cellcolor{nullCell}$-0.9$
& .689
\\

\bottomrule
\end{tabular}
\end{table}

\begin{figure}[t]
  \centering
\includegraphics[width=\textwidth]{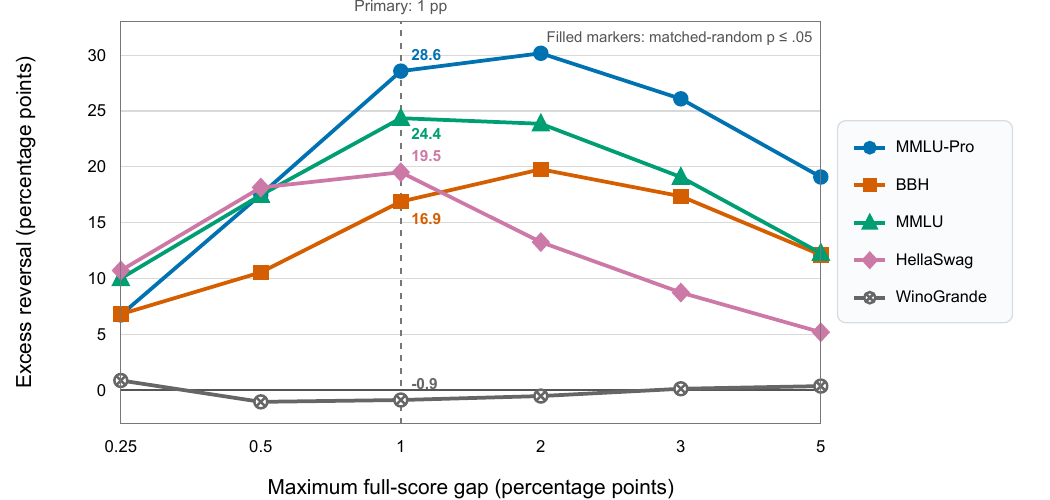}
  \caption{
Excess cross-family reversal rates over matched-random controls as the
maximum full-score gap defining a near tie is varied from 0.25 to 5
percentage points. The dashed line marks the pre-specified primary threshold
of 1 percentage point, and filled markers indicate one-sided matched-random
$p\leq .05$. Curves outside the primary threshold are descriptive sensitivity analyses.
}
  \label{fig:gaps}
\end{figure}

\paragraph{Generic and family-specific effects.}

Within-family near ties show excess reversals of 10.2--10.5 percentage points
in the four primary-positive benchmarks (all $p=.001$), indicating a generic
low-DIF recomposition component. The cross-family effect exceeds this
component by a further 6.7--18.0 points relative to paired matched-random
cross-minus-within gaps (all specificity $p=.001$). WinoGrande shows no
corresponding increment ($0.0$ points, $p=.496$;
Appendix~\ref{app:within-family-diagnostic}). Thus, the primary effect contains
both a generic recomposition component and an additional cross-family
component.

Despite this cross-family specificity, no family shows a consistent advantage
across benchmarks. Each family's mean anchor-minus-full score shift reverses
direction at least once. Exact-common-model comparisons confirm several such
reversals with the evaluated model population held fixed. The shifts are
therefore benchmark-dependent and should not be interpreted as intrinsic
family rankings.

\subsection{Robustness Analyses}
\label{sec:robustness-results}

\paragraph{Score-gap thresholds.}
Figure~\ref{fig:gaps} shows that the result is not specific to the
pre-specified 1-percentage-point near-tie threshold. The four
primary-positive benchmarks retain positive excess reversal rates across
wider thresholds, reaching 5.2--19.1 percentage points at 5 percentage
points, whereas WinoGrande remains near zero and is never significant.



\paragraph{Model population.}

\begin{figure}[t]
  \centering
  \includegraphics[width=\textwidth]{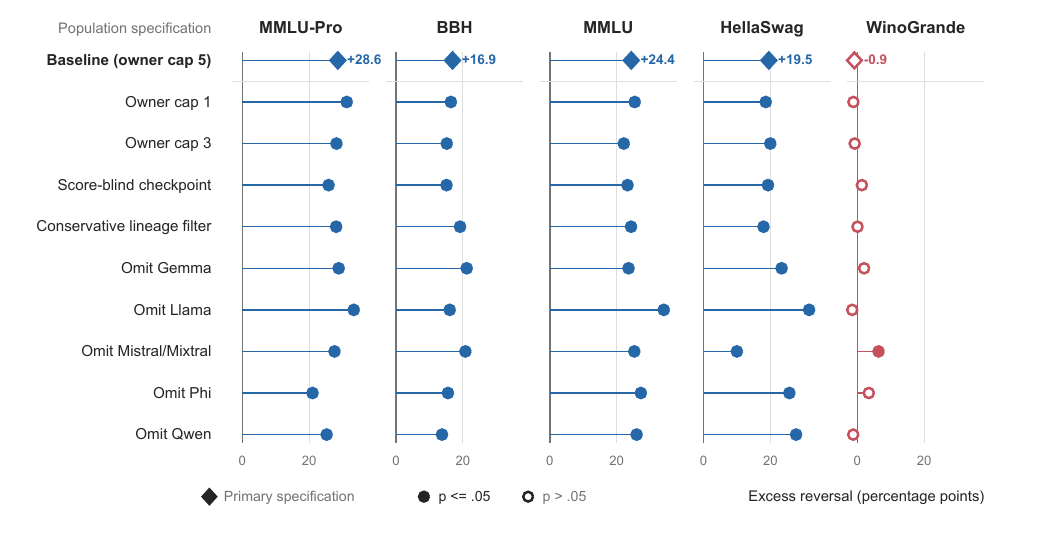}
  \caption{Population robustness of excess near-tie reversals.
Horizontal position gives the excess reversal rate in percentage
points. Diamonds mark the primary specification; filled markers denote
one-sided matched-random $p\leq.05$. The score-blind checkpoint variant
selects one model per owner--family pair by model-ID order without using
benchmark accuracy; conservative lineage filtering excludes name-flagged
merged, adapted, distilled, preference-tuned, hybrid, or uncensored
derivatives.}
  \label{fig:population-robustness}
\end{figure}

Across the baseline and all nine population perturbations described in
Section~\ref{sec:robustness}, MMLU-Pro, BBH, MMLU, and HellaSwag retain
positive excess reversal rates, with one-sided matched-random
$p\leq.05$ under every specification
(Figure~\ref{fig:population-robustness}). WinoGrande shows a significant
positive excess only when Mistral/Mixtral models are excluded.

\paragraph{Direct logistic MIRT.}

Replacing spectral ability adjustment with direct penalized joint logistic
MIRT preserves the four-benchmark pattern, but reduces excess reversals to
7.5--23.1 percentage points (all $p=.001$); WinoGrande remains nonsignificant
(2.7 points, $p=.087$). Thus, the qualitative primary finding persists, while
its magnitude depends on the adjustment estimator
(Appendix~\ref{app:direct-mirt}).

\subsection{Replication Across Owner Halves}
\begin{table}[t]
  \centering
  \caption{
    Owner-disjoint stability of residual item--family signatures.
    After adjustment within blueprint cells, $\rho$ is the median family-wise
    cross-half Spearman correlation. Top-20\% overlap measures discovery
    precision in the validation top 20\% (20\% under independence), and family
    agreement is computed conditional on overlapping top items. All 15
    one-sided within-cell permutation tests yield $p=.002$.
    }
  \label{tab:stability}
  \small
  \begin{tabular}{lrrr}
    \toprule
    Benchmark & Median $\rho$ & Top-20\% overlap & Advantaged-family agreement \\
    \midrule
    MMLU-Pro  & .407 & 31.5\% & 56.4\% \\
BBH       & .589 & 36.1\% & 72.5\% \\
MMLU      & .501 & 38.5\% & 71.9\% \\
HellaSwag & .308 & 37.4\% & 44.3\% \\
WinoGrande& .465 & 38.5\% & 66.0\% \\
    \bottomrule
  \end{tabular}
\end{table}

All three replication statistics for residual item--family signatures
show above-chance agreement across owner halves on every benchmark
(Table~\ref{tab:stability}).
Median family-wise correlations are
positive ($\rho=.308$--$.589$), while top-20\% overlap is 31.5--38.5\%,
compared with a 20\% independence baseline. Among overlapping top items, the
same advantaged family is recovered in 44.3--72.5\% of cases. Every statistic
exceeds all 500 corresponding within-cell permutation values (one-sided
$p=.002$; Bonferroni-adjusted $p=.030$ over 15 comparisons). WinoGrande remains
informative: its signatures replicate without excess near-tie reversals, so
replication alone does not imply a ranking effect.

Among the three benchmarks with multiple item groups,
discovery-selected item-group effect signs replicate in 58 of 60
held-out checks. Exact item-group contributions to the anchor-minus-full
family shifts retain their signs in 43 of 60 checks
(both pooled exact $p<.001$). These analyses localize reproducible
item-group structure but do not identify a cognitive or training-data
mechanism.

\subsection{Content Audit Results}
The blinded audit finds no confirmatory semantic enrichment.
Across the non-degenerate binary axes, the median Cohen's $\kappa$ is .715;
across the three ordinal axes, the median Spearman $\rho$ is .800, but none
of the ten pre-specified binary axes satisfies all dual-annotator confirmatory
criteria. The largest effect is annotator A's $-9.6$-percentage-point
spatial/temporal difference ($q=.106$); annotator B estimates $-0.8$ points
on the same axis ($q=.902$). Thus, the effect is neither significant after
correction nor replicated across annotators. No exploratory family-direction
association survives BH correction in either annotator.


\section{Limitations}



\paragraph{Scope and observational grouping.}
The five benchmarks are static, primarily closed-form, and drawn from
one response collection; MMLU and MMLU-Pro share content lineage and
are not independent replications
\citep{hendrycks2021mmlu,wang2024mmlupro}. Family labels are inferred
from model identifiers and may conflate architecture, training, data,
and uploader practices. Owner-disjoint evaluation and population
perturbations reduce specific artifacts but do not identify a causal
family mechanism. Low-DIF anchors do not define a superior measurement
construct or justify item removal, and the one-percentage-point
threshold is an operational choice rather than a universal decision
boundary.

\paragraph{Residual capability differences.}
Direct penalized joint logistic MIRT reproduces the qualitative
primary pattern, although effect sizes change
(Appendix~\ref{app:direct-mirt}).
Neither estimator establishes complete adjustment for capability
differences; nonlinear, higher-dimensional, or unmeasured structure
may remain. A post-hoc capability-profile overlap diagnostic finds limited
replacement coverage under stringent matching and does not establish
capability-equivalent subtests
(Appendix~\ref{app:capability_profile_overlap}).

\paragraph{Content audit.}
The two model annotators were not
calibrated against human expert labels, so their agreement may still reflect errors. The null findings on broad content axes therefore do not rule out finer-grained semantic explanations.

\section{Conclusion}

We audited whether cross-family leaderboard orderings separated by at most one
percentage point are robust to benchmark item composition. Although aggregate
rankings remain stable, four of five benchmarks show excess near-tie reversals
under residual low-DIF scoring; the fifth does not exceed matched-random
variation. Owner-disjoint estimation, matched controls, and population
robustness localize this sensitivity to near-tied comparisons rather than a
universal family advantage or broad leaderboard invalidity. Small score gaps
should therefore be accompanied by composition-robustness checks, not treated
as self-interpreting evidence of model superiority.

\clearpage

\medskip
\section*{AI use statement}

In this work, we used generative AI tools for assistance with code
implementation and literature searches, and as annotators in the
blinded content audit described in the paper. Additionally, we used
generative AI tools for formatting, language editing, and checks of
numerical consistency during manuscript preparation. We have reviewed
all AI-assisted work, including the implementation, cited references,
and manuscript revisions. We take responsibility for the final
content of this work, including text, claims, and artifacts produced
with the aid of generative AI.

\section*{Reproducibility Statement}
Section~\ref{sec:method} describes the data, model population, and owner-disjoint audit.
Appendices~\ref{app:mirt-implementation} and~\ref{app:validation-robustness}
provide implementation details for estimation, item selection, matched controls,
resampling, and validation.
Appendices~\ref{app:direct-mirt} and~\ref{app:capability_profile_overlap}
specify the direct logistic-MIRT comparison and capability-profile diagnostic,
respectively.
Appendix~\ref{app:parameter-rationale} explains the analysis settings and their limitations.

The code and reproduction instructions are publicly available at
\url{https://github.com/qiaoyuan667/near-tie-robustness}.
The repository includes data-preparation scripts that reconstruct the analysis
inputs from version-pinned upstream files with checksum verification.
It also provides the content-audit rubric, annotation schema, and numeric
annotation labels, allowing the reported annotation statistics to be
recomputed without new model calls.
Automated tests, numerical comparison tools, and a verification report
document the checks performed and their scope.
A command-line interface and worked example support applying the audit
to other benchmarks.
\bibliographystyle{iclr2027_conference}
\bibliography{references}

\appendix
\clearpage
\section{Comparison with the closest related work}
\label{app:related-work-comparison}

Table~\ref{tab:related-work} shows the comparison with the closest related work.

\begin{table}[!htbp]
\centering
\caption{
Comparison with the closest related work.
\cmark\ indicates direct coverage;
\tmark\ indicates partial or adjacent coverage;
\xmark\ indicates that the criterion is not studied.
}
\label{tab:related-work}

\normalsize

\setlength{\tabcolsep}{3pt}
\renewcommand{\arraystretch}{1.18}

\begin{tabularx}{\textwidth}{
@{}
>{\raggedright\arraybackslash}p{0.2\textwidth}
*{5}{>{\centering\arraybackslash}X}
@{}
}
\toprule
Work
& \shortstack{Ability-adjusted \\family DIF}
& \shortstack{Held-out\\validation}
& \shortstack{Item-set\\intervention}
& \shortstack{Matched\\controls}
& \shortstack{Near-tie\\replication}
\\
\midrule

\citet{liu2026bbhdif}
& \cmark & \cmark & \cmark & \xmark & \xmark \\

\citet{siska2024distribution}
& \xmark & \xmark & \cmark & \xmark & \tmark \\

\citet{halpin2024dtf}
& \tmark & \tmark & \cmark & \xmark & \xmark \\

BenchHub \citep{kim2026benchhub}
& \xmark & \xmark & \cmark & \xmark & \tmark \\

\citet{kotawala2026resolution}
& \xmark & \tmark & \xmark & \xmark & \tmark \\

Benchmark$^2$ \citep{qian2026benchmark2}
& \tmark & \cmark & \cmark & \xmark & \xmark \\

\citet{xu2026irtdrift}
& \tmark & \xmark & \xmark & \xmark & \xmark \\

\citet{habba2026growing}
& \tmark
& \cmark
& \cmark
& \tmark
& \xmark \\

\midrule
\textbf{Ours}
& \shortstack{\cmark}
& \shortstack{\cmark}
& \shortstack{\cmark}
& \shortstack{\cmark}
& \shortstack{\cmark}
\\
\bottomrule
\end{tabularx}
\end{table}

\section{Additional Results}

\subsection{Direct logistic MIRT comparison}
\label{app:direct-mirt}

\paragraph{Purpose and scope.}
We replace the spectral ability adjustment with direct penalized joint
logistic MIRT to test whether the primary ranking result depends on the
spectral approximation. This comparison fits item loadings and model
coordinates to the Bernoulli likelihood; it is not a marginal-likelihood or
Bayesian MIRT estimator and does not guarantee a globally optimal fit.
The response matrices, eligible model populations, owner-disjoint folds,
ranking representatives, inner splits, candidate dimensions, one-SE rule,
residual-DIF estimator, purification schedule, weighting, and resampling
procedures are unchanged. Selected dimensions and anchor identities are
re-estimated rather than held at their spectral values.

\paragraph{Likelihood and regularization.}
For a fitting matrix with $I$ items and $M$ models, let
$z_{im}=b_i+\mathbf a_i^\top\boldsymbol\theta_m$. We minimize
\begin{align}
\mathcal L
&=\frac{1}{IM}\sum_{i,m}
  \left[\log(1+e^{z_{im}})-y_{im}z_{im}\right] \nonumber\\
&\quad+\frac{\lambda}{2}\left(
  \frac{1}{I}\sum_i\|\mathbf a_i\|_2^2+
  \frac{1}{M}\sum_m\|\boldsymbol\theta_m\|_2^2\right)
  +\frac{\eta}{I}\sum_i b_i^2,
\label{eq:direct-mirt-objective}
\end{align}
with $\lambda=.01$ and $\eta=5\times10^{-5}$. The factor penalty is fixed
from the pilot rather than selected using outer ranking outcomes. Its scale
is not equivalent to the ridge penalty used to estimate held-out spectral
coordinates. We have not established robustness to this new penalty choice.

\paragraph{Fitting and dimension selection.}
Full-batch Adam uses learning rate .03, default moment parameters
$(.9,.999)$ and $\epsilon=10^{-8}$, and at most 6,000 updates. We check the
penalized training objective every 20 updates and stop after three consecutive
absolute changes below $10^{-6}$. Joint fits use two initializations and
retain the smaller returned penalized training objective. Loadings and
coordinates are initialized independently from a centered normal distribution
with standard deviation $0.1K^{-1/4}$; intercepts use
$\operatorname{logit}[(\sum_m y_{im}+.5)/(M+1)]$.
Joint and coordinate-only fits clip the gradient norm at 10. Computation uses
float32 on CPU with two threads. Selected fits must satisfy the recorded
stopping criterion, which indicates an objective plateau rather than a proof
of global convergence.

We use the same candidate grid
$K\in\{1,2,4,8,16,32,64,128,256\}$, omitting dimensions larger than
the smaller fitting-matrix dimension. Item parameters are learned using inner
training owners. With these parameters fixed, validation-model coordinates
are initialized at zero and fitted using calibration items only, with the
same per-model penalty. Disjoint evaluation items supply the held-out
Bernoulli log loss. As in the primary analysis, the smallest dimension within
one standard error of the minimum is selected; the standard error is computed
across validation-model mean losses, not independent owner clusters.

\paragraph{Purification and downstream scoring.}
In each purification round, model coordinates are learned only from the
currently selected items. These coordinates are centered and held fixed while
all item loadings and intercepts are fitted with the corresponding item
penalties in Equation~\ref{eq:direct-mirt-objective}. Selected-item parameters
provide a warm start; other loadings start at zero and their intercepts use
the smoothed empirical logit. The item-only fit uses the same Adam and plateau
settings, without gradient clipping. The resulting ability term
$\mathbf a_i^\top\boldsymbol\theta_m$, clipped to $[-6,6]$, replaces the
spectral offset in the unchanged residual-DIF routine. The MIRT intercept is
not added twice: the residual item intercept is re-estimated by that routine.
The original three purification rounds and final refit are retained.

\begin{table}[t]
\centering
\caption{Primary 50\% low-DIF results under direct penalized joint logistic
MIRT. Low-DIF and Random are reversal percentages for the same eligible
one-point pairs as Table~\ref{tab:main}; Random is the median of 1,000 matched
subtests. Excess and Spectral excess are in percentage points. $K$ gives
the two reselected dimensions. $p$ is the one-sided matched-random value.}
\label{tab:direct-mirt}
\small
\setlength{\tabcolsep}{4pt}
\begin{tabular}{lrrrrrr}
\toprule
Benchmark & $K$ & Low-DIF & Random & Excess & $p$ & Spectral excess \\
\midrule
MMLU-Pro & 32/32 & 41.6 & 18.5 & +23.1 & .001 & +28.6 \\
BBH & 128/128 & 32.7 & 25.2 & +7.5 & .001 & +16.9 \\
MMLU & 64/64 & 25.6 & 16.4 & +9.2 & .001 & +24.4 \\
HellaSwag & 16/16 & 23.4 & 11.5 & +12.0 & .001 & +19.5 \\
WinoGrande & 16/16 & 37.5 & 34.8 & +2.7 & .087 & $-0.9$ \\
\bottomrule
\end{tabular}
\end{table}

\paragraph{Results and interpretation.}
All five benchmarks and both outer folds completed, including the original
30\% and 70\% anchor-fraction analyses and primary resampling. At the primary
50\% fraction, the four originally positive benchmarks retain excess
reversals ($p=1/1001$ for each), whereas WinoGrande remains nonsignificant
($p=87/1001$; Table~\ref{tab:direct-mirt}). The four positive excesses decrease
under direct MIRT, especially on BBH and MMLU. Thus, the comparison supports
the qualitative primary pattern but not estimator-invariant effect sizes.
It does not establish that either representation removes every capability
difference or yields a better measurement construct.

Checks verify identical populations, representatives, and full-benchmark
scores; preservation of the total weight of every blueprint cell; and
primary rates and $p$-values reconstructed from pooled replicate counts. Population perturbations, wider-score-gap curves, discrimination and
reliability controls, pair-type specificity diagnostics, item-signature
stability, and semantic audits were not rerun with direct MIRT. Their reported
results remain specific to the primary spectral analysis.

\subsection{Discrimination- and reliability-matched diagnostic control}
\label{app:discrimination-control-results}

Table~\ref{tab:discrimination-matched-control} reports matching quality and
the resulting near-tie reversal rates.

\begin{table}[t]
\centering
\small
\caption{Discrimination-matched diagnostic control. Ratios compare the low-DIF
subtest with the median discrimination-matched control; values near one
indicate close balance. $|\Delta\alpha|$ is reported in units of $10^{-3}$.
``Rev.'' gives low-DIF/control reversal percentages.}
\label{tab:discrimination-matched-control}

\begin{tabular*}{\textwidth}{
@{\extracolsep{\fill}}
lrrrrrrr
@{}
}
\toprule
Benchmark &
$g_i$ ratio &
CITC ratio &
SEM ratio &
$10^3|\Delta\alpha|$ &
Rev.\ (\%) &
Excess (pp) &
$p$ \\
\midrule
MMLU-Pro   & .994  & .948 & .988  & 0.10 & 47.1/23.1 & +24.0 & .001 \\
BBH        & 1.001 & .962 & 1.003 & 0.42 & 42.1/28.0 & +14.2 & .001 \\
MMLU       & 1.000 & .959 & .994  & 0.11 & 40.7/19.4 & +21.3 & .001 \\
HellaSwag  & .982  & .960 & .966  & 0.14 & 30.9/12.9 & +18.0 & .001 \\
WinoGrande & 1.000 & .984 & 1.005 & 1.34 & 33.8/34.7 & $-0.8$ & .683 \\
\bottomrule
\end{tabular*}
\end{table}

The discrimination-matched controls achieve close balance with the low-DIF
subtests on all four measurement diagnostics. Across benchmarks, the
low-DIF-to-control ratio ranges from .982 to 1.001 for audit-half
predicted-logit spread, from .948 to .984 for corrected item--total
correlation, and from .966 to 1.005 for standard error of measurement; the
absolute Cronbach's-$\alpha$ difference is at most
$1.34\times10^{-3}$. Despite this balance, MMLU-Pro, BBH, MMLU, and
HellaSwag retain excess near-tie reversal rates of 14.2--24.0 percentage
points (all $p=.001$), whereas WinoGrande remains null
($-0.8$ points, $p=.683$). Thus, under these diagnostics, the primary result
is not explained by lower aggregate discrimination or reliability of the
low-DIF subtests.

\subsection{Within-family specificity diagnostic}
\label{app:within-family-diagnostic}

Table~\ref{tab:within-family-diagnostic} applies the frozen primary low-DIF
weights and the same 1,000 blueprint-matched random subtests to
different-owner, same-family near ties. ``Within L/R'' reports the low-DIF
reversal rate and the matched-random median. The specificity statistic compares
the observed cross-minus-within gap with the paired random-subtest
cross-minus-within distribution.

\begin{table}[t]
\caption{Post-hoc pair-type specificity diagnostic. Within $n$ is the pooled
number of different-owner, same-family pairs within one percentage point on
the full benchmark. Within L/R gives the low-DIF / matched-random reversal
rates. Within excess is their difference in percentage points.
$\Delta_{\mathrm{spec}}$ is the additional cross-family increment defined in
Equation~\ref{eq:specificity-excess}.}
\label{tab:within-family-diagnostic}
\centering
\small
\setlength{\tabcolsep}{3.8pt}
\begin{tabular}{lrrrrrr}
\toprule
Benchmark
& Within $n$
& Within L/R (\%)
& Within excess
& $p_{\mathrm{within}}$
& $\Delta_{\mathrm{spec}}$
& $p_{\mathrm{spec}}$ \\
\midrule
MMLU-Pro   & 256   & 27.0/16.4 & +10.5 & .001 & +18.0 & .001 \\
BBH        & 1,631 & 33.2/23.0 & +10.2 & .001 &  +6.7 & .001 \\
MMLU       & 4,523 & 24.1/13.8 & +10.3 & .001 & +13.9 & .001 \\
HellaSwag  & 5,199 & 20.4/10.2 & +10.2 & .001 &  +9.3 & .001 \\
WinoGrande & 5,409 & 30.9/31.7 & $-0.8$ & .723 & +0.0 & .496 \\
\bottomrule
\end{tabular}
\end{table}

The four primary-positive benchmarks therefore contain two components: a
generic within-family recomposition effect of approximately ten percentage
points and an additional cross-family increment of 6.7--18.0 points. All four
paired specificity tests attain the minimum possible empirical value
$p=1/1001=.001$. For WinoGrande, both pair types track their matched-random
baselines and the specificity contrast is null. These results do not imply
that every low-DIF-induced reversal is caused by family dependence; they show
that cross-family comparisons exhibit an additional component beyond generic
recomposition sensitivity.

\subsection{All score-gap thresholds}

\begin{table}
  \caption{Low-DIF reversal rate / matched-random median (\%) at cumulative full-score-gap thresholds. The frozen primary threshold is one percentage point.}
  \label{tab:gaps}
  \centering
  \small
  \resizebox{\textwidth}{!}{%
  \begin{tabular}{lrrrrrr}
    \toprule
    Benchmark & 0.25\pp & 0.5\pp & 1\pp & 2\pp & 3\pp & 5\pp \\
    \midrule
    MMLU-Pro  & 43.8/37.1 & 45.8/28.2 & 47.1/18.5 & 40.0/9.8 & 33.1/7.0 & 23.6/4.5 \\
BBH       & 49.1/42.3 & 46.6/36.0 & 42.1/25.2 & 33.9/14.1 & 26.9/9.6 & 18.0/5.9 \\
MMLU      & 48.0/38.0 & 46.3/28.8 & 40.7/16.3 & 31.7/7.8 & 24.1/5.1 & 15.3/3.1 \\
HellaSwag & 43.5/32.8 & 39.3/21.1 & 30.9/11.4 & 19.0/5.7 & 12.4/3.7 & 7.4/2.2 \\
WinoGrande& 46.0/45.1 & 40.5/41.5 & 33.8/34.7 & 22.6/23.1 & 16.4/16.3 & 10.6/10.3 \\
    \bottomrule
  \end{tabular}}
\end{table}

At the 0.25-point threshold, the eligible set falls to 89 pairs for MMLU-Pro,
and only BBH, MMLU, and HellaSwag pass the matched-random test.
Results outside the frozen one-point threshold are sensitivity analyses
rather than additional confirmatory tests.

Details are shown in Table~\ref{tab:gaps}.

\subsection{Anchor-fraction sensitivity}
\label{app:anchor-fraction-sensitivity}

Table~\ref{tab:anchor-fraction} reports descriptive sensitivity to the
fraction of low-DIF items retained within each blueprint cell.
The primary specification retains 50\% of items; 30\% and 70\% were
pre-specified sensitivity settings.

\begin{table}[t]
  \caption{Sensitivity to the retained anchor fraction. Each cell reports
  mean fold-specific Kendall $\tau_b$ / pooled strict near-tie reversal
  rate. Near-tied pairs are different-owner, different-family pairs within
  one percentage point on the full benchmark.}
  \label{tab:anchor-fraction}
  \centering
  \small
  \begin{tabular}{lccc}
    \toprule
    Benchmark & 30\% retained & 50\% retained & 70\% retained \\
    \midrule
    MMLU-Pro  & .885 / 48.4\% & .924 / 47.1\% & .949 / 39.3\% \\
BBH       & .882 / 44.3\% & .915 / 42.1\% & .951 / 33.9\% \\
MMLU      & .903 / 41.9\% & .930 / 40.7\% & .955 / 36.7\% \\
HellaSwag & .912 / 40.1\% & .948 / 30.9\% & .970 / 21.6\% \\
WinoGrande& .855 / 37.7\% & .900 / 33.8\% & .939 / 28.5\% \\
    \bottomrule
  \end{tabular}
\end{table}

As expected, retaining fewer items produces more ranking changes, but
near-tie reversals remain substantial across all three fractions. These
results are descriptive: matched-random controls were computed for the
primary 50\% specification only, so confirmatory excess-reversal inference
remains attached to that specification.

\subsection{Family/owner robustness matrix}
\label{sec:robustness_matrix}

\begin{table}
  \centering
  \caption{Excess one-point reversal rate, in percentage points, under the primary specification and nine frozen owner/family robustness variants. Stars indicate nominal randomization $p\leq.05$.}
  \label{tab:robust}
  \scriptsize
  \resizebox{\textwidth}{!}{%
  \begin{tabular}{lrrrrr}
    \toprule
    Variant & MMLU-Pro & BBH & MMLU & HellaSwag & WinoGrande \\
    \midrule
    Owner cap 1
  & +31.2$^{*}$ & +16.4$^{*}$ & +25.4$^{*}$ & +18.6$^{*}$ & $-1.2$ \\
Owner cap 3
  & +28.2$^{*}$ & +15.2$^{*}$ & +22.1$^{*}$ & +20.0$^{*}$ & $-0.8$ \\
Baseline cap 5
  & +28.6$^{*}$ & +16.9$^{*}$ & +24.4$^{*}$ & +19.5$^{*}$ & $-0.9$ \\
Score-blind checkpoint
  & +25.8$^{*}$ & +15.1$^{*}$ & +23.3$^{*}$ & +19.3$^{*}$ & +1.4 \\
Conservative lineage filter
  & +28.1$^{*}$ & +19.1$^{*}$ & +24.3$^{*}$ & +17.9$^{*}$ & +0.1 \\
Omit Gemma
  & +28.8$^{*}$ & +21.1$^{*}$ & +23.5$^{*}$ & +23.3$^{*}$ & +2.0 \\
Omit Llama
  & +33.3$^{*}$ & +16.1$^{*}$ & +34.1$^{*}$ & +31.6$^{*}$ & $-1.5$ \\
Omit Mistral/Mixtral
  & +27.6$^{*}$ & +20.7$^{*}$ & +25.3$^{*}$ & +10.0$^{*}$ & +6.4$^{*}$ \\
Omit Phi
  & +21.0$^{*}$ & +15.5$^{*}$ & +27.2$^{*}$ & +25.7$^{*}$ & +3.4 \\
Omit Qwen
  & +25.2$^{*}$ & +13.7$^{*}$ & +26.0$^{*}$ & +27.7$^{*}$ & $-1.2$ \\
    \bottomrule
  \end{tabular}}
\end{table}

The baseline row uses 1,000 matched-random controls, so its minimum attainable randomization $p$-value is $1/1001$; each of the other nine variants uses 200 controls, giving a minimum of $1/201$. Under the frozen robustness rule, a specification passes the direction criteria when at least three of the five benchmarks have positive excess and passes the significance criteria when at least three have nominal $p\leq.05$. All ten specifications pass both criteria. More strongly, MMLU-Pro, BBH, MMLU, and HellaSwag---the four primary-positive benchmarks---remain positive and nominally significant in every specification. WinoGrande is significant only under omission of Mistral/Mixtral, so it remains an unstable boundary case rather than a robust positive result. The details are shown in Table~\ref{tab:robust}.

\subsection{Exact-common-model directional checks}
\label{sec:exact_common_direction}

\begin{table}
  \centering
  \caption{Three illustrative changes in family mean anchor-minus-full score shifts across benchmarks, evaluated on exactly the same family-specific models in each benchmark pair. Shifts and differences are reported in percentage points; 95\% confidence intervals are obtained from 2,000 owner-bootstrap replicates.}
  \label{tab:direction}
  \small
  \resizebox{\textwidth}{!}{%
  \begin{tabular}{llrrrr}
    \toprule
    Exact-common pair & Family & Left shift & Right shift & Right$-$left & [95\% CI] \\
    \midrule
    MMLU / HellaSwag
  & Gemma & +2.04 & $-0.51$ & $-2.55$ & [$-3.07$, $-1.91$] \\
HellaSwag / WinoGrande
  & Llama & $-1.48$ & +0.14 & +1.63 & [+1.37, +1.88] \\
MMLU-Pro / BBH
  & Gemma & $-0.28$ & +0.82 & +1.10 & [+0.38, +1.94] \\
    \bottomrule
  \end{tabular}}
\end{table}

Table~\ref{tab:direction} reports three compact examples used in the main text rather than an exhaustive search for the largest contrasts. Each comparison holds the family-specific model population fixed across the two benchmarks, ruling out changes in model membership as the explanation for the directional change. The released file contains all 15 family estimates for these three
exact-common benchmark pairs, including sample sizes and confidence
intervals.
These shifts are descriptive consequences of benchmark composition and should not be interpreted as intrinsic or causal family rankings.

\subsection{Blinded content-audit results}
\label{sec:blinded_audit}

Using the 250 exact-cell matched pairs and two blinded annotators defined in
Appendix~\ref{app:content-audit}, Figure~\ref{fig:content-audit} reports the
pooled paired prevalence differences for the ten prespecified confirmatory
binary axes. Annotation reliability passes the frozen criteria. Across the
non-degenerate binary axes, the median Cohen's $\kappa$ is .715; across the
three ordinal axes, the median Spearman $\rho$ is .800.

No binary axis passes the dual-annotator confirmatory enrichment criteria. The
largest absolute difference for annotator A is $-9.6$ percentage points on
spatial/temporal content ($q=.106$), while annotator B estimates $-0.8$ points
on the same axis ($q=.902$). The largest absolute difference for annotator B is
+4.0 points on distractor discrimination ($q=.902$). Thus, no axis is both
significant after correction and replicated across annotators.

A post-confirmatory diagnostic is restricted to the 158 of 250
high-residual-DIF items for which the advantaged-family label agrees across
owner halves. No benchmark-stratified family-direction association survives
BH correction in either annotator, and no dual-annotator exploratory signal is
detected. These analyses remain exploratory and are not used as a semantic
explanation of the ranking effect.

\begin{figure}[t]
  \centering
  \includegraphics[width=\textwidth]{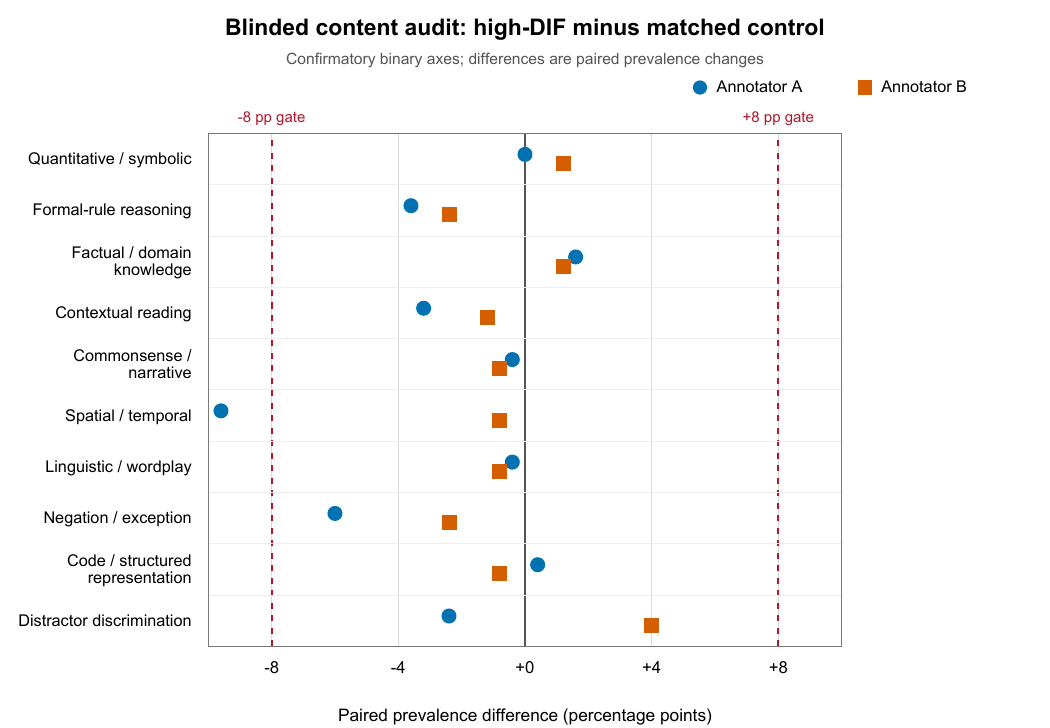}
  \caption{Paired prevalence differences between stable high-DIF items and
exact-cell matched controls on the ten confirmatory binary content axes.
Dashed lines mark the frozen 8-percentage-point effect-size requirement.
Annotator A crosses the magnitude threshold only for spatial/temporal content
($-9.6$ points), but the effect is not significant after BH correction
($q=.106$) and is not replicated by annotator B ($-0.8$ points). No axis
passes the dual-annotator confirmatory criteria.}
  \label{fig:content-audit}
\end{figure}

\section{Implementation details for the MIRT-adjusted audit}
\label{app:mirt-implementation}

This appendix specifies the numerical implementation underlying
Section~\ref{sec:method}. The main text defines the statistical estimand;
here we describe the spectral MIRT working representation, dimension selection,
residual-DIF optimization, anchor purification, and resampling procedures.

\subsection{Model-family and owner construction}
\label{app:population-construction}

Family membership is assigned by a unique, case-insensitive match on model ID.
The frozen matching rules are:

\begin{center}
\begin{tabular}{ll}
\toprule
Family & Model-ID pattern \\
\midrule
Qwen & \texttt{qwen} \\
Llama & \texttt{llama} \\
Gemma & \texttt{gemma} \\
Mistral/Mixtral & \texttt{mistral} or \texttt{mixtral} \\
Phi & a token beginning with \texttt{phi} \\
\bottomrule
\end{tabular}
\end{center}

Models matching zero or multiple family patterns are excluded. The owner is the
uploader component of the model ID. If an owner contributes more than the
frozen cap of five checkpoints to one family, five are selected using the
frozen population-selection seed.

Owners are ordered by decreasing retained-checkpoint count, with ties broken
by a seeded random key. Each entire owner is then assigned to the half that
minimizes the sum of the normalized within-family checkpoint-count imbalance
and the within-family full-accuracy residual imbalance; exact cost ties are
resolved using the frozen population-selection seed. No owner may appear in
both halves.

In benchmark order (MMLU-Pro, BBH, MMLU, HellaSwag, and WinoGrande), the
complete-response matrices contain respectively 12,032, 5,761, 14,042,
10,042, and 1,267 items and 1,823, 3,811, 5,000, 5,000, and 5,000 models.
The corresponding ranking populations contain 204, 497, 673, 673, and 673
owner--family representatives.

\subsection{Spectral working-response construction}
\label{app:spectral-mirt}

Let $\mathcal A$ denote the models in the current audit half and let
$M_{\mathcal A}=|\mathcal A|$. For every item, we calculate a smoothed marginal
success probability
\begin{equation}
  \widehat p_i
  =
  \frac{
    \sum_{m\in\mathcal A}Y_{im}+1/2
  }{
    M_{\mathcal A}+1
  },
  \qquad
  s_i
  =
  \sqrt{\widehat p_i(1-\widehat p_i)}.
  \label{eq:app-smoothed-probability}
\end{equation}
We then form the standardized working-response matrix
\begin{equation}
  X_{im}
  =
  \frac{Y_{im}-\widehat p_i}{s_i}.
  \label{eq:app-working-response}
\end{equation}

Suppose $\mathcal S$ is the item set currently used to estimate the model
coordinates. We compute a rank-$K$ randomized singular-value decomposition
\citep{halko2011randomized}
\begin{equation}
  \mathbf X_{\mathcal S}
  \approx
  \mathbf U_K\mathbf\Sigma_K\mathbf V_K^\top.
  \label{eq:app-svd}
\end{equation}
The model-coordinate matrix is
\begin{equation}
  \mathbf\Theta=\mathbf V_K,
  \label{eq:app-model-coordinates}
\end{equation}
and loadings for all items, including those outside $\mathcal S$, are obtained
by projection:
\begin{equation}
  \mathbf L
  =
  \mathbf X\mathbf\Theta.
  \label{eq:app-item-loadings}
\end{equation}
Thus the rank-$K$ reconstruction of the standardized response residual is
$\mathbf L\mathbf\Theta^\top$. We convert it into the fixed offset used in the
logistic residual-DIF model:
\begin{equation}
  o_{im}
  =
  \operatorname{clip}
  \left(
    \frac{
      \mathbf L_{i,:}\mathbf\Theta_{m,:}^{\top}
    }{s_i},
    -6,6
  \right).
  \label{eq:app-offset}
\end{equation}
The truncation at six logits is a numerical safeguard and is not tuned using
family labels or ranking outcomes.

In the numerical implementation, $o_{im}$ is the capability term
$\eta_{im}^{(K)}$ used in Section~\ref{sec:method}.
Its dependence on the dimension $K$, the current audit half, and the
fitting item set $\mathcal S$ is suppressed in the notation.
It contains only the scaled and clipped spectral contribution, not an
item intercept or a residual family effect. The offset is held fixed
within each residual-DIF fit, but is recomputed between purification
rounds.

This construction is a spectral approximation to a multidimensional logistic
response surface. It is not a full maximum-likelihood calibration of a
traditional 2PL MIRT model; we therefore refer to it as a spectral MIRT working
model.

\subsection{Nested selection of the MIRT dimension}
\label{app:mirt-dimension}

For outer audit half $h$, let $M_{\mathrm{train},h}$ denote the
number of models belonging to the inner-training owners. The fitting
matrix has $I$ item rows and $M_{\mathrm{train},h}$ model columns,
so the feasible candidate set is
\[
\mathcal{K}_h =
\left\{
K\in\{1,2,4,8,16,32,64,128,256\}:
K\leq \min\{I,M_{\mathrm{train},h}\}
\right\}.
\]
The feasible grid reaches $K=128$ in both MMLU-Pro folds and $K=256$ in
all folds of the other four benchmarks.
Dimension selection is performed separately inside each outer audit half.

We permute the unique audit-half owner IDs using the frozen inner-split seed
and assign the first rounded 25\% to inner validation, bounded to leave at
least one owner on each side. All models from an owner receive the same
assignment. Using only
the remaining owners, we compute Equation~\ref{eq:app-working-response} and a
rank-$K$ decomposition. For dimension $K$, define the training-owner item
factor matrix
\begin{equation}
  \mathbf F_K
  =
  \mathbf U_K\mathbf\Sigma_K.
  \label{eq:app-item-factors}
\end{equation}

Within each metadata-defined item group, items are randomly divided
between a coordinate-calibration set $\mathcal C$ and an evaluation set
$\mathcal E$, with approximately half assigned to each. For
inner-validation model $m$, its coordinates are estimated only from
$\mathcal C$:
\begin{equation}
  \widehat{\boldsymbol\theta}_{m,K}
  =
  \left(
    \mathbf F_{\mathcal C,K}^{\top}
    \mathbf F_{\mathcal C,K}
    +
    \lambda_\theta\mathbf I_K
  \right)^{-1}
  \mathbf F_{\mathcal C,K}^{\top}
  \mathbf X_{\mathcal C,m},
  \qquad
  \lambda_\theta=1.
  \label{eq:app-coordinate-ridge}
\end{equation}

For evaluation item $i\in\mathcal E$ and inner-validation model $m$,
the full predicted logit used for dimension selection is
\begin{equation}
  \widehat{\xi}^{\mathrm{CV}}_{im,K}
  =
  \operatorname{logit}(\widehat p_i)
  +
  \operatorname{clip}
  \left(
    \frac{
      \mathbf F_{i,K}^{\top}
      \widehat{\boldsymbol\theta}_{m,K}
    }{s_i},
    -6,6
  \right).
  \label{eq:app-cv-prediction}
\end{equation}
Here $\widehat p_i$ and $s_i$ are computed from the inner-training
owners. Unlike the capability offset in Equation~\ref{eq:app-offset},
$\widehat{\xi}^{\mathrm{CV}}_{im,K}$ includes the item baseline
$\operatorname{logit}(\widehat p_i)$; it contains no residual family
effect.

The model-level validation loss is
\begin{equation}
  L_{m,K}
  =
  \frac{1}{|\mathcal E|}
  \sum_{i\in\mathcal E}
  \left[
    \log\left(
      1+\exp\left(\widehat{\xi}^{\mathrm{CV}}_{im,K}\right)
    \right)
    -
    Y_{im}\widehat{\xi}^{\mathrm{CV}}_{im,K}
  \right].
  \label{eq:app-cv-loss}
\end{equation}

Let $\overline L_K$ be the mean of $L_{m,K}$ over inner-validation
models, and define
\begin{equation}
\operatorname{SE}_K
=
\frac{
\operatorname{sd}_m(L_{m,K})
}{
\sqrt{M_{\mathrm{val}}}
}.
\end{equation}
Let
$K_{\mathrm{best}}
=
\arg\min_{K\in\mathcal{K}_h} \overline{L}_K$.
Following the one-standard-error rule, we select
\begin{equation}
K^\star
=
\min\left\{
K\in\mathcal{K}_h:
\overline{L}_K\leq \overline{L}_{K_{\mathrm{best}}}+\mathrm{SE}_{K_{\mathrm{best}}}
\right\}.
\label{eq:dimension-selection}
\end{equation}
The selected $K^\star$ is frozen before family labels are introduced and is
used for every anchor fraction within that outer fold. 

The one-SE selection remains below the feasible candidate ceiling in every
fold. The raw held-out-loss minimum occurs at the feasible ceiling in five of
the ten folds, so the selected representation should not be interpreted as
exhausting all possible capability structure.

\subsection{Residual-DIF optimization}
\label{app:dif-optimization}

With the spectral capability offset $o_{im}$ held fixed at the
fold-selected dimension $K^\star$, we estimate the item intercepts
and residual family effects using the response logit
\begin{equation}
  \xi_{im}
  =
  b_i+o_{im}+\delta_{i,f(m)}.
  \label{eq:app-dif-logit}
\end{equation}
This is the numerical implementation of the main-text predictor,
with $o_{im}$ corresponding to $\eta_{im}^{(K^\star)}$.
The item intercept $b_i$ is re-estimated by the optimization below;
the empirical baseline $\operatorname{logit}(\widehat p_i)$ used
for inner validation in Equation~\ref{eq:app-cv-prediction} is not
added as a separate term here.

Let $\mathcal A$ denote the models in the current fitting owner half
and $M_{\mathcal A}=|\mathcal A|$. We initialize
\begin{align}
\widetilde p_i
&=
\operatorname{clip}\!\left(
\frac{1}{M_{\mathcal A}}
\sum_{m\in\mathcal A}Y_{im},
10^{-4},
1-10^{-4}
\right),\\
b_i^{(0)}
&=
\operatorname{logit}(\widetilde p_i)
-
\frac{1}{M_{\mathcal A}}
\sum_{m\in\mathcal A}o_{im},
\label{eq:app-intercept-initialization}
\end{align}
and set all family effects to zero. Here
$\operatorname{clip}(x,\ell,u)=\max\{\ell,\min(x,u)\}$.

At the current parameter values, let
\[
p_{im}=\sigma(\xi_{im}).
\]
For fixed family effects, the score and information for item
intercept $b_i$ are
\begin{align}
g_{b_i}
&=
\sum_m(Y_{im}-p_{im}),\\
H_{b_i}
&=
\sum_m p_{im}(1-p_{im}).
\end{align}
We apply the clipped Newton update
\begin{equation}
b_i
\leftarrow
b_i+
\operatorname{clip}\!\left(
\frac{g_{b_i}}{H_{b_i}+10^{-9}},
-2,2
\right).
\label{eq:app-intercept-update}
\end{equation}

For family $f$, let
\[
\mathcal M_f=\{m:f(m)=f\}.
\]
Holding the item intercept fixed, the penalized score and information
for $\delta_{if}$ are
\begin{align}
g_{\delta_{if}}
&=
\sum_{m\in\mathcal M_f}(Y_{im}-p_{im})
-\lambda\delta_{if},\\
H_{\delta_{if}}
&=
\sum_{m\in\mathcal M_f}p_{im}(1-p_{im})
+\lambda,
\end{align}
with $\lambda=1$. We update
\begin{equation}
\delta_{if}
\leftarrow
\delta_{if}
+
\operatorname{clip}\!\left(
\frac{g_{\delta_{if}}}{H_{\delta_{if}}},
-2,2
\right).
\label{eq:app-family-update}
\end{equation}
The probabilities $p_{im}$ are recomputed at the current parameter
values during each Newton iteration.

After every complete family-effect sweep, we impose the identifying
constraint by setting
\begin{equation}
\overline\delta_i
=
\frac{1}{F}\sum_f\delta_{if},
\qquad
\delta_{if}
\leftarrow
\delta_{if}-\overline\delta_i,
\qquad
b_i
\leftarrow
b_i+\overline\delta_i.
\label{eq:app-family-centering}
\end{equation}
The simultaneous adjustment of $b_i$ preserves every linear
predictor while ensuring
$\sum_f\delta_{if}=0$.

We use at most five alternating intercept/family-effect cycles.
Each Newton subproblem uses at most 30 iterations and terminates when
the largest absolute update is below $10^{-8}$. The outer alternation
also terminates early when the largest family-effect update is below
this threshold.

\subsection{Anchor purification algorithm}
\label{app:purification}

For each outer fold and anchor fraction $\alpha$, purification proceeds as
follows.

\begin{enumerate}
  \item Initialize
  \[
    \mathcal S^{(0)}=\{1,\ldots,I\}.
  \]

  \item For rounds $t=1,2,3$:

  \begin{enumerate}
    \item Fit the rank-$K^\star$ spectral MIRT representation using the response
    rows in $\mathcal S^{(t-1)}$.

    \item Project all items onto the fitted model coordinates and construct
    offsets $o_{im}^{(t)}$.

    \item Fit $b_i^{(t)}$ and $\delta_{if}^{(t)}$ on all items conditional on
    $o_{im}^{(t)}$.

    \item Compute
    \[
      D_i^{(t)}
      =
      \max_f\delta_{if}^{(t)}
      -
      \min_f\delta_{if}^{(t)}.
    \]

    \item Within each item group $s$ containing $n_s$ items, order items by
$b_i^{(t)}$, breaking ties by original item index, and divide them into
$\min(4,n_s)$ nonempty, near-equal strata. Each resulting subset forms
a blueprint cell.

    \item Within every blueprint cell $c$, retain
    \[
      k_c
      =
      \max\{1,\operatorname{round}(\alpha n_c)\}
    \]
    items with the smallest $D_i^{(t)}$. Their union defines
    $\mathcal S^{(t)}$.
  \end{enumerate}

  \item Assign final weights
  \[
    w_i
    =
    \begin{cases}
      n_c/k_c, & i\in\mathcal S^{(3)}\cap c,\\
      0,       & \text{otherwise}.
    \end{cases}
  \]

  \item Refit the spectral representation and residual-DIF parameters once on
  $\mathcal S^{(3)}$ to obtain the stored final item signatures. Ranking scores
  use the weights selected at the end of round three.
\end{enumerate}

Purification always runs for three rounds; convergence of the anchor set is
recorded through successive Jaccard overlap but is not used as a stopping rule.

\subsection{Matched-random subtests}
\label{app:random-control}

For each outer fold, let $\mathcal S_c$ be the final low-DIF anchors in
blueprint cell $c$ and $k_c=|\mathcal S_c|$. Random-control replicate $r$ independently samples
\begin{equation}
\mathcal{S}_c^{(r)}
\subseteq
\{i : c(i)=c\},
\qquad
\left|\mathcal{S}_c^{(r)}\right|=k_c,
\label{eq:app-random-sample}
\end{equation}
uniformly without replacement. Sampled items receive weight $n_c/k_c$.

The same replicate index is pooled across the two outer target halves:
\begin{equation}
  T_r
  =
  \frac{
    \sum_h N_{\mathrm{flip},h}^{(r)}
  }{
    \sum_h N_{\mathrm{eligible},h}^{(r)}
  }.
  \label{eq:app-pooled-control}
\end{equation}
The reported matched-random median, interval, excess, and randomization
$p$-value are calculated from $\{T_r\}_{r=1}^{1000}$.

An additional diagnostic control that also matches discrimination and
reliability is defined in
Appendix~\ref{app:discrimination-control-method}.

\subsection{Composition bootstrap}
\label{app:composition-bootstrap}

For benchmarks with at least five metadata-defined item groups, each
item group serves as a bootstrap unit. Otherwise, the original item
order is divided deterministically into 100 contiguous, near-equal
clusters, which serve as the bootstrap units.

Let $B$ be the number of bootstrap units and let
\begin{equation}
\left(
N_1^{(r)},\ldots,N_B^{(r)}
\right)
\sim
\operatorname{Multinomial}
\left(
B;
\frac{1}{B},\ldots,\frac{1}{B}
\right).
\label{eq:app-bootstrap-multiplicities}
\end{equation}

be the unit multiplicities for replicate $r$.

For unit $b$, let $S_{bm}$ and $n_b$ be the model's full-score successes and
item mass, and let $\widetilde S_{bm}$ and $\widetilde n_b$ be their weighted
anchor equivalents. Bootstrap scores are
\begin{equation}
  A_m^{(r)}
  =
  \frac{
    \sum_bN_b^{(r)}S_{bm}
  }{
    \sum_bN_b^{(r)}n_b
  },
  \qquad
  \widetilde A_m^{(r)}
  =
  \frac{
    \sum_bN_b^{(r)}\widetilde S_{bm}
  }{
    \sum_bN_b^{(r)}\widetilde n_b
  }.
  \label{eq:app-bootstrap-scores}
\end{equation}
The close-pair mask is defined once from the observed full scores and held fixed
across the 2,000 bootstrap replicates.

\begin{table}[!htb]
\centering
\caption{Composition-bootstrap sensitivity under the primary
50\% low-DIF specification and one-percentage-point near-tie
threshold. Ranges are the 2.5th--97.5th percentiles across
2,000 item-group/block-bootstrap replicates.
For each replicate, reversal counts and eligible-pair counts
are pooled across the two target halves before calculating
the reversal rate; Kendall correlations are averaged across
halves. These are ranges of recomposed statistics, not
confidence intervals around the fixed-benchmark estimates
or intervals for excess reversals.}
\label{tab:composition-bootstrap-ranges}
\begin{tabular}{lccc}
\toprule
Benchmark &
Observed reversal (\%) &
Bootstrap range (\%) &
Bootstrap $\tau_b$ range \\
\midrule
MMLU-Pro  & 47.1 & 31.0--51.9 & .913--.933 \\
BBH       & 42.1 & 18.9--39.2 & .870--.928 \\
MMLU      & 40.7 & 31.1--45.5 & .905--.936 \\
HellaSwag & 30.9 & 28.3--33.6 & .939--.949 \\
WinoGrande & 33.8 & 23.4--34.9 & .839--.884 \\
\bottomrule
\end{tabular}
\end{table}

\subsection{Implementation settings}
\label{app:frozen-settings}

Table~\ref{tab:v3-hyperparameters} summarizes the frozen primary-audit
specification.

\begin{table}[!htb]
  \centering
  \caption{Implementation settings for the primary spectral audit.}
  \label{tab:v3-hyperparameters}
  \small
  \begin{tabularx}{\linewidth}{
    @{}
    >{\raggedright\arraybackslash}p{0.29\linewidth}
    >{\raggedright\arraybackslash}X
    @{}
  }
    \toprule
    Component & Frozen Setting \\
    \midrule
    Families
      & Qwen, Llama, Gemma, Mistral/Mixtral, Phi \\

    Owner cap
      & 5 checkpoints per owner--family \\

    Minimum capped family size
      & 20 models \\

    Outer split
      & owner-disjoint two-fold cross-fit \\

    MIRT candidate dimensions
      & $\{1,2,4,8,16,32,64,128,256\}$;\newline
        values exceeding the smaller fitting-matrix dimension omitted \\

    Selected $K$ (two folds)
      & MMLU-Pro 64/32; BBH 128/128; MMLU 128/128;\newline
        HellaSwag 16/64; WinoGrande 8/8 \\

    Dimension rule
      & minimum $K$ within one SE of best loss \\
    Inner validation owners & 25\% \\
    Coordinate-calibration items & 50\% within each item group \\
    Coordinate ridge $\lambda_\theta$ & 1 \\
    Maximum MIRT offset & 6 logits \\
    Randomized SVD iterations & 5 \\
    DIF penalty $\lambda_\delta$ & 1 \\
    DIF fitting cycles & 5 \\
    Newton iterations per update & at most 30 \\
    Newton tolerance & $10^{-8}$ \\
    Maximum Newton step & 2 logits \\
    Purification rounds & 3 \\
    Easiness strata & Up to 4 within each item group \\
    Primary anchor fraction & 50\% \\
    Anchor sensitivity & 30\%, 70\% \\
    Primary close-pair band & 1 percentage point \\
    Gap sensitivity & 0.25, 0.5, 1, 2, 3, 5 points \\
    Matched-random replicates & 1,000 \\
    Composition-bootstrap replicates & 2,000 \\
    Single-group bootstrap units & 100 contiguous clusters \\
    \bottomrule
  \end{tabularx}
\end{table}

\section{Validation and Robustness: Implementation Details}
\label{app:validation-robustness}

\subsection{Controls and uncertainty}
\label{app:controls-uncertainty}

\paragraph{Eligible pairs and pooled reversal rate.}
Let $h\in\{D,V\}$ index the target owner half and let $a$ denote an alternative
scoring rule, either the low-DIF anchor score or one matched-random score. For
threshold $\epsilon$, define
\begin{equation}
\mathcal{P}_{h}^{(a)}(\epsilon)
=
\left\{
(m,m') :
\begin{array}{l}
m<m',\quad
o_m\neq o_{m'},\quad
f_m\neq f_{m'},\\[2pt]
0<
\left|
s_m^{\mathrm{full}}-s_{m'}^{\mathrm{full}}
\right|
\leq\epsilon,\quad
s_{mh}^{(a)}\neq s_{m'h}^{(a)}
\end{array}
\right\},
\label{eq:eligible-pair-set}
\end{equation}
where $o_m$ and $f_m$ denote the owner and family of model $m$. Thus,
eligible models must come from different owners and different families, must
be separated by no more than $\epsilon$ on the full benchmark, and must not
be tied under either scoring rule.

For an eligible pair, the strict reversal indicator is
\begin{equation}
R_{hmm'}^{(a)}
=
\mathbf{1}\!\left[
\left(
s_m^{\mathrm{full}}-s_{m'}^{\mathrm{full}}
\right)
\left(
s_{mh}^{(a)}-s_{m'h}^{(a)}
\right)
<0
\right].
\label{eq:strict-reversal-indicator}
\end{equation}
The pooled reversal rate sums reversal counts and eligible-pair counts across
the two cross-fitting directions:
\begin{equation}
\widehat{T}_{\epsilon}^{(a)}
=
\frac{
\displaystyle
\sum_h
\sum_{(m,m')\in\mathcal{P}_{h}^{(a)}(\epsilon)}
R_{hmm'}^{(a)}
}{
\displaystyle
\sum_h
\left|
\mathcal{P}_{h}^{(a)}(\epsilon)
\right|
}.
\label{eq:pooled-reversal-rate}
\end{equation}
The primary analysis sets $\epsilon=0.01$, corresponding to a
one-percentage-point full-benchmark score gap.

\paragraph{Resampling controls.}
Appendix~\ref{app:random-control} specifies the matched-random construction
and fold pooling, while Appendix~\ref{app:composition-bootstrap} specifies
the composition resampling. The present subsection defines eligible pairs
and the additional threshold and owner-bootstrap analyses.

\paragraph{Discrimination- and reliability-matched diagnostic control.}
\label{app:discrimination-control-method}

Within each outer fold, let $\mathcal{M}_{\mathrm{audit}}$ denote
the models in the audit owner half, corresponding to $\mathcal A$
in Appendix~\ref{app:spectral-mirt}.
For item $i$, let $o_{im}$ be the audit-half capability offset
defined in Equation~\ref{eq:app-offset}, evaluated at the
fold-selected dimension $K^\star$.
We define the spectral discrimination proxy as its standard
deviation across audit-half models:
\begin{equation}
  g_i
  =
  \operatorname{SD}_{m\in\mathcal{M}_{\mathrm{audit}}}
  \left(o_{im}\right).
  \label{eq:audit-logit-spread}
\end{equation}
This diagnostic summarizes across-model variation in the spectral
capability term, without adding the item intercept or residual
family effect. It uses the audit-half offsets, not the
inner-validation predictions in
Equation~\ref{eq:app-cv-prediction}.
Target-half responses do not enter its construction.

Within each frozen blueprint cell $c$, we order items by $g_i$,
breaking ties by original item index, and divide them into
\begin{equation}
q_c=\min\{4,n_c\}
\end{equation}
nonempty, near-equal discrimination strata, where $n_c$ is the number of
items in the original cell. Within each refined stratum, each control replicate samples
without replacement exactly the number of items contributed by the frozen
low-DIF subtest. If the low-DIF subtest contains $k_c$ items in the original blueprint cell $c$, every sampled control item in that cell receives the
same restoring weight $n_c/k_c$. Thus, every control matches the low-DIF
subtest's total length and exact selected-item counts across the refined
strata while preserving the original total weight of every blueprint cell. We generate 1,000 such controls independently in each audit fold.

All matching variables are estimated using the audit owner half. The selected
control items and their weights are then frozen before application to the
opposite, owner-disjoint target half. Target-half responses are used only for
ranking and for held-out balance diagnostics; they do not affect control
selection.

For target-half models $\mathcal{M}_{\mathrm{target}}$, the corrected
item--total correlation of item $i$ is
\begin{equation}
r_i
=
\operatorname{Corr}_{m\in\mathcal{M}_{\mathrm{target}}}
\left(
Y_{im},
\sum_{j\neq i}Y_{jm}
\right).
\label{eq:heldout-citc}
\end{equation}
For a subtest with nonnegative item weights $\mathbf{w}$, we summarize held-out
item discrimination by the weighted mean
\begin{equation}
\overline{r}(\mathbf{w})
=
\frac{\sum_i w_i r_i}{\sum_i w_i}.
\label{eq:weighted-citc}
\end{equation}

Let
\begin{equation}
s_i^2
=
\operatorname{Var}_{m\in\mathcal{M}_{\mathrm{target}}}(Y_{im})
\end{equation}
denote the held-out response variance of item $i$. Under the independent-item
approximation, the standard error of the normalized weighted score is
\begin{equation}
\operatorname{SEM}(\mathbf{w})
=
\frac{
\sqrt{\sum_i w_i^2s_i^2}
}{
\sum_i w_i
}.
\label{eq:independent-item-sem}
\end{equation}
This is used as a comparative score-precision diagnostic rather than as a
claim that benchmark-item responses are statistically independent.

For target model $m$, define the unnormalized weighted total
\begin{equation}
S_m(\mathbf{w})=\sum_i w_iY_{im},
\end{equation}
and let
\begin{equation}
K_{\mathbf{w}}=\left|\{i:w_i>0\}\right|
\end{equation}
be the number of selected items. The weighted Cronbach's $\alpha$ diagnostic is
\begin{equation}
\alpha(\mathbf{w})
=
\frac{K_{\mathbf{w}}}{K_{\mathbf{w}}-1}
\left[
1-
\frac{
\sum_i w_i^2s_i^2
}{
\operatorname{Var}_{m\in\mathcal{M}_{\mathrm{target}}}
\left(S_m(\mathbf{w})\right)
}
\right].
\label{eq:weighted-cronbach-alpha}
\end{equation}

For each benchmark, low-DIF diagnostic values are averaged across the two
owner-disjoint scoring directions. For every control-replicate index, the
corresponding diagnostic is likewise averaged across directions, and the
median over 1,000 replicates forms the control reference. The reported
$g_i$, CITC, and SEM ratios divide the low-DIF value by this median; the
reported $|\Delta\alpha|$ is the absolute difference between the low-DIF
$\alpha$ and the median control $\alpha$. Reversal counts and eligible-pair
counts are pooled across the two directions exactly as in the primary
matched-random analysis. Excess reversals and the one-sided randomization
$p$-value are then computed relative to the 1,000
discrimination-matched replicates.

\paragraph{Post-hoc pair-type specificity diagnostic.}
Let
$\widehat{T}_{\epsilon,\mathrm{cross}}^{(a)}$
denote the pooled reversal rate in
Equation~\ref{eq:pooled-reversal-rate} under the primary different-family
restriction. We define
$\widehat{T}_{\epsilon,\mathrm{within}}^{(a)}$
analogously by replacing $f_m\neq f_{m'}$ in
Equation~\ref{eq:eligible-pair-set} with $f_m=f_{m'}$, while retaining the
different-owner restriction, the frozen full-score gap, and
scoring-rule-specific tie exclusions.

The within-family excess reversal rate is
\begin{equation}
\Delta_{\mathrm{within}}
=
100\left[
\widehat{T}_{0.01,\mathrm{within}}^{(\mathrm{DIF})}
-
\operatorname{median}_{1\leq r\leq R}
\widehat{T}_{0.01,\mathrm{within}}^{(r)}
\right],
\label{eq:within-family-excess}
\end{equation}
with one-sided empirical value
\begin{equation}
p_{\mathrm{within}}
=
\frac{
1+
\sum_{r=1}^{R}
\mathbf{1}\!\left[
\widehat{T}_{0.01,\mathrm{within}}^{(r)}
\geq
\widehat{T}_{0.01,\mathrm{within}}^{(\mathrm{DIF})}
\right]
}{
R+1
}.
\label{eq:within-family-p}
\end{equation}

To test whether the cross-family effect exceeds the generic within-family
effect, define
\begin{equation}
G_{\mathrm{DIF}}
=
\widehat{T}_{0.01,\mathrm{cross}}^{(\mathrm{DIF})}
-
\widehat{T}_{0.01,\mathrm{within}}^{(\mathrm{DIF})},
\qquad
G_r
=
\widehat{T}_{0.01,\mathrm{cross}}^{(r)}
-
\widehat{T}_{0.01,\mathrm{within}}^{(r)}.
\label{eq:pair-type-gap}
\end{equation}
The family-specificity excess is
\begin{equation}
\Delta_{\mathrm{spec}}
=
100\left[
G_{\mathrm{DIF}}
-
\operatorname{median}_{1\leq r\leq R}G_r
\right],
\label{eq:specificity-excess}
\end{equation}
and its one-sided paired empirical value is
\begin{equation}
p_{\mathrm{spec}}
=
\frac{
1+
\sum_{r=1}^{R}
\mathbf{1}\!\left[
G_r\geq G_{\mathrm{DIF}}
\right]
}{
R+1
}.
\label{eq:specificity-p}
\end{equation}
The low-DIF and random cross-minus-within gaps use the same frozen model
population and the same random-subtest replicate index. Because this
comparison was added after the primary analysis, it is interpreted as a
post-hoc diagnostic rather than a confirmatory estimand.

\paragraph{Threshold sensitivity.}
The primary specification uses a 50\% anchor fraction and defines near-tied
models as those separated by at most one percentage point on the full
benchmark. We additionally evaluate
\begin{equation}
\epsilon
\in
\{0.0025,0.005,0.01,0.02,0.03,0.05\},
\end{equation}
corresponding to full-score gaps of
$\{0.25,0.5,1,2,3,5\}$ percentage points. Pair membership at every threshold
is frozen from the observed full-benchmark scores. The matched-random
distribution is recomputed using 1,000 replicates, and composition uncertainty
is recomputed using 2,000 bootstrap replicates. We also repeat anchor
construction using the lowest-DIF 30\% and 70\% of items within each blueprint
cell. These specifications are sensitivity analyses and are not used to
select the primary threshold or anchor fraction.

\paragraph{Owner bootstrap.}
Family-specific score and percentile shifts may be correlated within a model
owner. We therefore resample owners rather than checkpoints. Let
\begin{equation}
d_m=s_m^{\mathrm{anchor}}-s_m^{\mathrm{full}}
\end{equation}
be the score shift for representative model $m$. In bootstrap replicate $b$,
we sample the observed owners with replacement, using a sample size equal to
the number of unique owners, and include all representative records associated
with each sampled owner. For family $f$, the bootstrap mean is
\begin{equation}
\widehat{\mu}_{f}^{(b)}
=
\frac{1}{|\mathcal{M}_{f}^{(b)}|}
\sum_{m\in\mathcal{M}_{f}^{(b)}}d_m.
\end{equation}
We analogously calculate the mean percentile-rank shift and report empirical
95\% intervals over 2,000 replicates. Owner-bootstrap intervals are used only
for family-specific shifts; uncertainty in the primary reversal statistic is
handled by the composition bootstrap and matched-random control.

\subsection{Population robustness}
\label{app:population-robustness}

The primary population retains at most five checkpoints per
owner--family and selects one ranking representative nearest the
within-owner--family median full-benchmark accuracy. Entire owners,
including all of their family-specific records, remain assigned to a
single outer half. We evaluate nine perturbations of this
construction:

\begin{enumerate}
    \item reduce the owner cap from five checkpoints to one;
    \item reduce the owner cap from five checkpoints to three;
    \item select one checkpoint per owner--family lexicographically by model
          identifier, without using benchmark accuracy;
    \item exclude model identifiers indicating merged, adapter-based,
          distilled, hybrid, preference-tuned, or uncensored lineages; and
    \item omit each of Gemma, Llama, Mistral/Mixtral, Phi, and Qwen in turn.
\end{enumerate}

The conservative lineage filter excludes identifiers matching terms including
\texttt{merge}, \texttt{slerp}, \texttt{franken}, \texttt{lora},
\texttt{qlora}, \texttt{adapter}, \texttt{distill}, \texttt{hybrid},
\texttt{fusion}, \texttt{dpo}, \texttt{ppo}, \texttt{orpo}, \texttt{kto},
\texttt{abliterat}, and \texttt{uncensor}. Each retained family must contain at
least 15 models in these robustness populations. Owner-disjoint discovery and
validation halves are then reconstructed using the frozen population seed.

For every perturbation, the spectral MIRT dimension $K$ is fixed to the value
selected for the corresponding benchmark and audit fold in the primary
population. We refit the MIRT representation, residual family-DIF model, and
three-round anchor-purification procedure under the perturbed population, but
do not reselect $K$. All remaining parameters are held at their primary
values, including the 50\% anchor fraction and one-percentage-point close-pair
threshold.

The owner-cap-five baseline retains its original 1,000 matched-random
replicates. Each new perturbation uses 200 matched-random replicates. For
benchmark $b$ and perturbation $v$, define
$\widehat{\Delta}_{bv}$ and $p_{bv}$ as the matched-random excess and
one-sided randomization $p$-value. The prespecified robustness counts are
\begin{align}
G_v^{\mathrm{dir}}
&=
\sum_{b=1}^{5}
\mathbf{1}\!\left[\widehat{\Delta}_{bv}>0\right],\\
G_v^{\mathrm{sig}}
&=
\sum_{b=1}^{5}
\mathbf{1}\!\left[p_{bv}\leq0.05\right].
\end{align}
A perturbation passes when
\begin{equation}
G_v^{\mathrm{dir}}\geq3
\qquad\text{and}\qquad
G_v^{\mathrm{sig}}\geq3.
\label{eq:population-robustness-gate}
\end{equation}
The population-selection audit, including family counts, owner counts, and
cross-fitting-half counts, was written before the perturbed DIF analyses were
executed.

\subsection{Item-signature replication and item-group attribution}
\label{app:item-stability-source}

\paragraph{Common blueprint cells.}
Item-signature replication is evaluated by refitting the residual family-DIF
model independently in the two owner-disjoint halves. These analyses reuse the
owner-cap-five baseline population and its frozen owner split. Let
$b_i^{(D)}$ and $b_i^{(V)}$ denote the item intercepts estimated in the
discovery and validation halves. We define the common item intercept
\begin{equation}
\bar b_i
=
\frac{b_i^{(D)}+b_i^{(V)}}{2}
\end{equation}
and assign each item to a common blueprint cell
\begin{equation}
c(i)
=
\left(
\operatorname{group}(i),
Q_4(\bar b_i)
\right),
\label{eq:common-blueprint-cell}
\end{equation}
where $\operatorname{group}(i)$ denotes item $i$'s metadata-defined item
group and $Q_4$ assigns its easiness stratum within that group using
$\bar b_i$. Because the same cells are used in both halves, stability
cannot be induced by independently changing the stratification boundaries.

\paragraph{Family-effect replication.}
Let $\delta_{if}^{(h)}$ be the residual effect of family $f$ on item $i$ in
owner half $h\in\{D,V\}$. Before calculating correlations, we remove the mean
effect within every common cell:
\begin{equation}
\widetilde{\delta}_{if}^{(h)}
=
\delta_{if}^{(h)}
-
\frac{1}{|c(i)|}
\sum_{j:c(j)=c(i)}
\delta_{jf}^{(h)}.
\end{equation}
For each family, we calculate the Spearman rank correlation
\citep{spearman1904proof}
\begin{equation}
\rho_f
=
\rho_S\left(
\widetilde{\delta}^{(D)}_{\cdot f},
\widetilde{\delta}^{(V)}_{\cdot f}
\right)
\end{equation}
and use $\operatorname{median}_f\rho_f$ as the benchmark-level replication
statistic.

Within every common cell, we separately identify the top 20\% of items by DIF
magnitude in each owner half. Let $H_D$ and $H_V$ denote these sets. We report
the discovery-to-validation overlap precision
\begin{equation}
O
=
\frac{|H_D\cap H_V|}{|H_D|}
\end{equation}
and, among stable high-DIF items, the agreement in the maximally advantaged
family:
\begin{equation}
A
=
\frac{1}{|H_D\cap H_V|}
\sum_{i\in H_D\cap H_V}
\mathbf{1}
\left[
\arg\max_f\delta_{if}^{(D)}
=
\arg\max_f\delta_{if}^{(V)}
\right].
\end{equation}

The null distribution is constructed from 500 permutations. In each
replicate, the complete validation-half item signature is permuted among items
within the same common blueprint cell. For statistic $S$, the one-sided
permutation $p$-value is
\begin{equation}
p_{\mathrm{perm}}(S)
=
\frac{
1+\sum_{r=1}^{500}
\mathbf{1}[S^{(r)}\geq S^{\mathrm{obs}}]
}{
501
}.
\end{equation}

\paragraph{Interpretation relative to chance.}
We interpret these statistics using their null behavior rather than additional
study-defined cutoffs. Under no cross-half association, the family-wise
Spearman correlations are centered at zero, while independent top-20\%
selections have expected overlap precision 20\%. The chance level of
advantaged-family agreement depends on the empirical family frequencies and
is therefore obtained from the within-cell permutation distribution. Across
the five benchmarks and three statistics, all 15 observed values exceed all
500 corresponding permutation values:
\[
p_{\mathrm{perm}}=\frac{1}{501}=.002.
\]
Even a Bonferroni correction over the 15 comparisons gives
$p_{\mathrm{adj}}=.030$.

\paragraph{Family effects within item groups.}
Item-group attribution is restricted to MMLU-Pro, BBH, and MMLU, which
contain multiple metadata-defined item groups. To remove global offsets,
each family's item effects are centered by that family's all-item mean
separately in each owner half:
\begin{equation}
\delta_{if}^{*(h)}
=
\delta_{if}^{(h)}
-
\frac{1}{I}\sum_{j=1}^{I}\delta_{jf}^{(h)}.
\end{equation}
For item group $s$, let $I_s$ denote its set of items. The mean family
effect within that group is
\begin{equation}
\gamma_{sf}^{(h)}
=
\frac{1}{|I_s|}
\sum_{i\in I_s}\delta_{if}^{*(h)}.
\end{equation}
Only item groups containing at least 20 items are eligible. For each
family, the two most positive and two most negative group effects in
the discovery half are selected, and their signs are tested in the
validation half. Pooled support requires at least 70\% sign replication
with a one-sided exact binomial $p\leq0.05$, together with at least 65\%
replication in at least two of the three benchmarks included in this
analysis.

\paragraph{Exact item-group contributions to score shifts.}
As a secondary attribution analysis, we exactly decompose each
cross-fitted family score shift into contributions from individual
item groups. If anchors fitted in half $h$ are evaluated on target
half $\bar h$, the contribution of item group $s$ to family $f$ is
\begin{equation}
C_{sf}^{(h\rightarrow\bar h)}
=
\frac{100}{
I\,|\mathcal{M}_{f,\bar h}|
}
\sum_{m\in\mathcal{M}_{f,\bar h}}
\sum_{i\in I_s}
\left(w_i^{(h)}-1\right)Y_{im}.
\label{eq:source-contribution}
\end{equation}
Because the anchor weights sum to $I$,
\begin{equation}
\sum_s C_{sf}^{(h\rightarrow\bar h)}
=
100\left(
\bar s_{f}^{\mathrm{anchor}}
-
\bar s_{f}^{\mathrm{full}}
\right)
\end{equation}
up to numerical precision. For each family, the two largest positive and two
largest negative contributions in the discovery-to-validation direction are
selected, and their signs are checked after swapping the two folds. This decomposition is a secondary, mechanism-adjacent analysis and is not used
to establish owner-disjoint item-signature replication.

\subsection{Blinded content-audit protocol and statistical tests}
\label{app:content-audit}

\paragraph{Matched item sample.}
The content audit is conducted only after establishing that item-family
signatures replicate across owner halves. Within every benchmark, stable
high-DIF items are defined as items that fall in the top 20\% of DIF magnitude
within their common blueprint cell in both owner halves. Controls must fall
outside the top 20\% in both halves. We sample 50 stable high-DIF items per benchmark and match each one without
replacement to a control from the same common blueprint cell. This produces 50 matched pairs, or 100 items,
per benchmark and 500 items in total.

Each item is assigned a random blinded identifier. Annotators receive only
this identifier and the question text. Benchmark, item-group label, matched-pair identity, experimental group,
family effects, advantaged family, and DIF magnitudes are stored
separately and are not included in the annotation prompt.

\paragraph{Annotation procedure.}
Annotator A used GPT-5.5 \citep{openai2026gpt55}, and Annotator B used GPT-5.4
\citep{openai2026gpt54}; each annotated all 500 items.
Questions are processed in batches of 20 with low reasoning effort. The
annotation protocol and output schema were frozen before either annotation
run. Annotators are not asked to solve the benchmark items or predict which
models answer them correctly. Instead, they label visible content and
reasoning demands along ten binary axes:
\begin{quote}
quantitative or symbolic reasoning; formal rule reasoning; factual domain
knowledge; contextual reading; commonsense or narrative reasoning;
spatial--temporal reasoning; linguistic wordplay; negation or exception
handling; code or structured representation; and distractor discrimination.
\end{quote}
They additionally assign ordinal scores from 0 to 3 for reasoning steps,
context burden, and ambiguity. The complete rubric and structured annotation
schema are included in the released artifact.

\paragraph{Reliability criteria.}
For each non-degenerate binary axis, agreement between annotators is
measured using Cohen's $\kappa$ \citep{cohen1960coefficient}. For each
ordinal axis, agreement is measured using Spearman's $\rho$
\citep{spearman1904proof}. Annotation reliability passes when
\begin{equation}
\operatorname{median}_{j}\kappa_j \geq 0.45
\qquad\text{and}\qquad
\operatorname{median}_{k}\rho_k \geq 0.50.
\label{eq:annotation-reliability-gate}
\end{equation}
The median for binary axes excludes axes for which $\kappa$ is undefined
because both annotators assign a constant label.

\paragraph{Paired semantic tests.}
Each annotator is analyzed separately; annotations are not averaged or
adjudicated. For binary axis $j$ and matched pair $p$, define
\begin{equation}
d_{pj}
=
X_{pj}^{\mathrm{high}}
-
X_{pj}^{\mathrm{control}}.
\end{equation}
The pooled prevalence difference is
\begin{equation}
\widehat d_j
=
\frac{1}{250}\sum_{p=1}^{250}d_{pj}.
\end{equation}
We apply the exact two-sided McNemar test
\citep{mcnemar1947sampling}, equivalently an exact binomial test on the
two types of discordant pairs. Benjamini--Hochberg correction
\citep{benjamini1995controlling} is applied over the ten binary axes
separately for each annotator. Ordinal axes are evaluated with paired
Wilcoxon signed-rank tests \citep{wilcoxon1945individual} and separately
corrected, but are not included in the confirmatory semantic criteria.

A binary axis provides a replicated content explanation only if all of the
following conditions hold:

\begin{enumerate}
    \item the pooled difference has the same nonzero direction for both
          annotators;
    \item the BH-adjusted value satisfies $q\leq0.05$ for both annotators;
    \item $|\widehat d_j|\geq0.08$ for both annotators; and
    \item the pooled direction occurs in at least three of the five benchmarks
          for both annotators.
\end{enumerate}

The confirmatory content audit passes if at least one prespecified binary axis
meets all four requirements.

\paragraph{Post-confirmatory family analysis.}
Among the 250 stable high-residual-DIF items, 158 receive the same
advantaged-family label in both owner halves. For each annotator,
associations between these stable labels and the ten binary content axes
are evaluated using benchmark-stratified label-permutation tests with
10,000 permutations, followed by Benjamini--Hochberg correction across
axes. Because this analysis followed the confirmatory audit, it is
exploratory and is not part of the semantic criteria.

\section{Reproducibility and artifact boundary}
\label{sec:reproducibility}

The public repository contains the analysis code, final experimental
protocols, aggregate outputs, figure-generation code, and automated validation
tests. The tests cover model-family assignment, owner-disjoint splitting,
representative selection, family-blind spectral-MIRT dimension selection,
residual-DIF estimation, cross-fitted anchor construction, matched-random
controls, score-gap sensitivity, population robustness, item-signature stability, item-group attribution, and blinded
content-audit analysis.
\section{Rationale and limitations of analysis settings}
\label{app:parameter-rationale}

Our settings serve distinct roles: some define the estimand, some control
sample construction, and others regularize or numerically approximate a fit.
A fixed setting is not thereby optimal, and a pre-specified decision threshold
is not a universal standard of measurement quality. We distinguish the
rationale for a choice from sensitivity evidence supporting it.

\paragraph{Population and estimand.}
The five named families and unique-match assignment define an observational
comparison rather than a taxonomy of all models. Capping each owner--family
at five limits checkpoint proliferation while retaining within-owner
variation. Caps of one and three, score-blind representative selection,
conservative lineage filtering, and family omissions test specific alternatives.
The minimum family sizes of 20 in the primary analysis and 15 in population
perturbations are operational sample-support cutoffs, not power guarantees.
Two owner-disjoint folds allow each half to select items for the other without
sharing owners. Median-accuracy representatives avoid selecting the best
checkpoint; the model-ID alternative tests dependence on that rule.
The one-percentage-point band defines the near-tie estimand. The .25--5-point
analysis assesses other bands without redefining the primary endpoint.

\paragraph{Ability adjustment and regularization.}
The powers-of-two grid spans coarse to flexible linear representations while
limiting the number of fits. The one-SE rule favors a smaller representation
with comparable held-out prediction \citep{hastie2009elements}; it does not
establish complete capability adjustment. Holding out 25\% of inner owners reserves data for selection while keeping
most owners for fitting. Splitting items approximately in half within
each item group separates coordinate calibration from predictive evaluation. These proportions were not separately optimized.
Spectral coordinate ridge 1 stabilizes calibration, and residual-DIF penalty 1
shrinks poorly supported family effects. Their numerical values remain
implementation choices, not empirically established optima. Direct MIRT adds
a differently scaled factor penalty and optimization settings
(Appendix~\ref{app:direct-mirt}); the primary comparison tests an estimator
change, not a comprehensive regularization sweep.

\paragraph{Item selection and weighting.}
Using up to four easiness strata within each item group provides a coarse
balance between difficulty control and cell size. This stratification
does not guarantee matched item discrimination or test information. Selecting 50\% low-DIF items trades targeted filtering
against retained test length; 30\% and 70\% provide fraction sensitivity.
At least one selected item per nonempty cell keeps cell-restoring weights
defined, and those weights preserve the original cell mass. Three
purification rounds bound repeated fitting; the final refit estimates
signatures for the frozen selection. Neither three rounds nor four easiness
bins has been shown to be optimal.

\paragraph{Numerical and Monte Carlo settings.}
Smoothed empirical probabilities, probability clipping, bounded offsets,
limited Newton steps, and denominator floors prevent undefined logits,
unstable updates, or extreme extrapolation. Five randomized-SVD power
iterations and the fixed residual-fitting schedule bound computation;
the Newton tolerance controls individual updates rather than proving global
convergence. These safeguards are specified in
Appendices~\ref{app:frozen-settings} and~\ref{app:direct-mirt}.
With the plus-one randomization formula \citep{phipson2010permutation},
1,000 primary random subtests provide minimum attainable $p=1/1001$;
200 per population perturbation provide $1/201$; and 500 signature
permutations provide $1/501$. These counts set Monte Carlo resolution, not
effect size. The 2,000 bootstrap resamples and 95\% percentile intervals
provide a fixed uncertainty summary. When fewer than five metadata-defined item groups are available,
resampling 100 contiguous item blocks provides a proxy for dependence
among items. These blocks are not assumed to be natural or independent
clusters.

\paragraph{Replication and semantic interpretation.}
Top-20\% sets focus replication on pronounced residual signatures while
retaining enough items for comparison. The 20\% overlap reference assumes
independent uniform selection; within-cell permutations provide the
design-specific reference for the reported tests. Item-group attribution requires at least 20 items per group and selects
the two most positive and two most negative group effects for each family.
These choices limit small-group noise and the number of directional
checks. The directional cutoffs in
Appendix~\ref{app:item-stability-source} are operational summaries;
checks that share families or item groups are not independent replications.

The content audit samples 50 matched pairs per benchmark to give equal
benchmark representation within a finite annotation budget. Ten binary axes
and three ordinal axes define a deliberately coarse taxonomy. Two annotators
permit an agreement check but do not replace human validation. Reliability
cutoffs and the requirement of an eight-point effect, matching directions,
and evidence in at least three benchmarks are conservative operational
criteria; they are not justified by a formal power calculation. In the pooled
250-pair audit, eight percentage points correspond to 20 net discordant pairs.
BH correction controls multiplicity within the stated families of tests
\citep{benjamini1995controlling}. The 10,000-permutation family-direction
analysis is exploratory. Neither passing a reliability threshold nor failing
the enrichment criteria establishes semantic validity or absence of content
effects.

\section{Capability-profile overlap diagnostic}
\label{app:capability_profile_overlap}

\paragraph{Purpose and status.}
This exploratory diagnostic was specified after the primary analysis, with its
protocol fixed before the diagnostic runs. It asks whether the available item
pool supports replacing low-DIF items with alternatives that have similar full
estimated capability profiles. We reuse the primary spectral analysis's frozen
owner-disjoint folds, selected dimensions, 50\% low-DIF items, blueprint cells,
scoring weights, and one-percentage-point near-tie definition. This is not a
reanalysis of the direct logistic MIRT variant or a confirmatory test of a causal
family mechanism.

\paragraph{Audit-only profiles and matching.}
For an audit half with $M_A$ models, let $L_{i,:}$ be item $i$'s full $K$-dimensional
spectral loading, $s_i$ its working-response Bernoulli scale, and $b_i$ its fitted
item intercept. Define
\[
E_i = \frac{L_{i,:}}{s_i\sqrt{M_A}},
\qquad
 d_{\mathrm{prof}}(i,j) = \lVert E_i-E_j\rVert_2.
\]
Because the spectral model coordinates have orthonormal columns,
$d_{\mathrm{prof}}$ equals the root-mean-square difference between the two items'
predicted ability offsets across audit models, before logit clipping.
All $K$ coordinates participate, preserving both direction and magnitude rather
than matching only a scalar discrimination proxy.
Within each frozen blueprint cell, we match selected low-DIF items to nonselected
items without replacement, using cost
\[
c_{ij}=d_{\mathrm{prof}}(i,j)^2+(b_i-b_j)^2.
\]
Unconstrained matching minimizes total cost. Caliper matching requires both
$d_{\mathrm{prof}}(i,j)\leq\tau$ and $|b_i-b_j|\leq\tau$, first maximizing the
number of matched pairs and then minimizing cost.
We evaluate unconstrained matching and $\tau\in\{0.25,0.5,1.0\}$ logits, together
with a nonoptimized random-pairing reference: five scenarios per benchmark.
These calipers are diagnostic settings, not established standards of acceptable
balance. No target-half responses enter the profiles, pairings, or caliper choices.

\paragraph{Paired controls and distinct contrasts.}
Each of 1,000 controls independently selects either endpoint of every pair with
probability $1/2$ and retains unmatched low-DIF items. Each selected endpoint
receives its cell's frozen positive restoring weight $n_c/k_c$, preserving every
cell's selected-item count and total scoring weight, without duplicate items. Reversal and eligible-pair counts are
pooled across target folds before calculating rates, retaining the primary
scoring-rule-specific tie exclusions.
The paired-mixture excess compares the low-DIF reversal rate with the median
control rate. Its one-sided empirical tail value uses the primary plus-one
Monte Carlo formula.
Under unconstrained matching, a paired-mixture control overlaps the original
low-DIF subtest by approximately 50\%; it is not a high-DIF-only subtest.
Nonselected partners are not assumed to have strictly larger final-refit DIF.

A separate all-swapped contrast replaces every matched low-DIF item with its
partner. We compare the low-DIF-minus-all-swapped reversal difference with the
differences from complementary random endpoint configurations, using a two-sided
empirical reference. These two contrasts answer different questions.
All reported $p$-values are unadjusted exploratory tail values conditional on the
constructed pairings, not causal tests based on randomized family dependence.

\begin{table}[!htbp]
\centering
\small
\setlength{\tabcolsep}{3pt}
\caption{Unconstrained profile-aware matching. Low, Mix, and Swap are reversal
percentages for the low-DIF subtest, median paired-mixture control, and all-swapped
subtest, respectively. $\Delta_{\mathrm{mix}}$ is Low minus the mixture median;
$\Delta_{\mathrm{swap}}$ is Low minus Swap, both in percentage points.
$p_{\mathrm{mix}}$ is one-sided and $p_{\mathrm{swap}}$ is two-sided.
Differences use unrounded values; $.001$ displays $1/1001$.}
\label{tab:profile_overlap_unconstrained}
\begin{tabular}{lrrrrrrr}
\toprule
& & \multicolumn{3}{c}{Paired mixture} & \multicolumn{3}{c}{All-swapped contrast} \\
\cmidrule(lr){3-5}\cmidrule(lr){6-8}
Benchmark & Low & Mix & $\Delta_{\mathrm{mix}}$ & $p_{\mathrm{mix}}$
& Swap & $\Delta_{\mathrm{swap}}$ & $p_{\mathrm{swap}}$ \\
\midrule
MMLU-Pro  & 47.1 & 17.5 & $+29.5$ & .001 & 39.9 & $+7.1$ & .066 \\
BBH       & 42.1 & 22.9 & $+19.2$ & .001 & 36.8 & $+5.3$ & .003 \\
MMLU      & 40.7 & 14.7 & $+26.0$ & .001 & 43.0 & $-2.3$ & .039 \\
HellaSwag & 30.9 & 10.6 & $+20.3$ & .001 & 24.5 & $+6.4$ & .001 \\
WinoGrande& 33.8 & 33.1 & $+0.7$  & .297 & 37.1 & $-3.2$ & .086 \\
\bottomrule
\end{tabular}
\end{table}

\paragraph{Unconstrained results do not establish balance.}
The four primary-positive benchmarks retain excess reversals relative to the
profile-aware paired mixtures, whereas WinoGrande remains nonsignificant
(Table~\ref{tab:profile_overlap_unconstrained}). However, minimum total matching
cost does not imply small absolute distances.
Table~\ref{tab:profile_overlap_balance} shows substantial remaining profile and
predicted-probability differences. In particular, the first three benchmarks have
probability RMS differences of 26.0--27.5 percentage points. HellaSwag also retains
a large continuous-intercept difference despite within-cell matching.
Thus, these pairings do not establish capability-equivalent subtests. Small
imbalances in vector means would not establish pairwise or joint-distribution
balance either.
The all-swapped contrasts also vary in sign and nominal significance; the
paired-mixture result must not be restated as a uniform low-DIF-versus-high-DIF
ordering.

\begin{table}[!htbp]
\centering
\small
\setlength{\tabcolsep}{5pt}
\caption{Unconstrained matching quality. Each entry averages the two fold-specific
medians over matched pairs. Profile RMS and $|\Delta b|$ are in logits; probability
RMS is in percentage points and excludes the fitted family-effect term.
Cosine similarity measures profile direction rather than magnitude.}
\label{tab:profile_overlap_balance}
\begin{tabular}{lrrrr}
\toprule
Benchmark & Profile RMS & Cosine & $|\Delta b|$ & Probability RMS \\
\midrule
MMLU-Pro   & 1.917 & .547 & .409  & 26.8 \\
BBH        & 1.686 & .640 & .248  & 26.0 \\
MMLU       & 1.811 & .594 & .294  & 27.5 \\
HellaSwag  & 2.306 & .753 & 1.359 & 9.9  \\
WinoGrande & .706  & .918 & .270  & 9.3  \\
\bottomrule
\end{tabular}
\end{table}

\paragraph{Calipers expose limited replacement support.}
Coverage denotes the fraction of original low-DIF scoring weight belonging to
matched items, not the fraction actually replaced in a random control; the latter
is one-half of coverage in expectation.
At $\tau=0.5$, coverage is negligible for MMLU-Pro, BBH, and MMLU
(Table~\ref{tab:profile_overlap_caliper}). MMLU-Pro has only 28 matched pairs across
both folds, and MMLU has one pair with identical audit- and target-half responses.
The corresponding controls are therefore almost unchanged low-DIF subtests.
Their nonsignificant results do not establish that the effect disappears under
successful, broad capability matching.

\begin{table}[!htbp]
\centering
\small
\setlength{\tabcolsep}{6pt}
\caption{Caliper diagnostic with both profile RMS and $|\Delta b|$ bounded by
$0.5$ logits. Coverage is replaceable low-DIF scoring weight, in percent.
Excess is relative to paired-mixture controls, in percentage points; displayed
zeroes are rounded. Low coverage limits the informativeness of the contrast.}
\label{tab:profile_overlap_caliper}
\begin{tabular}{lrrr}
\toprule
Benchmark & Coverage (\%) & Excess (pp) & $p$ \\
\midrule
MMLU-Pro   & 0.233  & 0.0    & .955  \\
BBH        & 0.103  & 0.0    & .611  \\
MMLU       & 0.007  & 0.0    & 1.000 \\
HellaSwag  & 13.152 & $+0.5$ & .089  \\
WinoGrande & 30.844 & $-0.6$ & .798  \\
\bottomrule
\end{tabular}
\end{table}

At $\tau=1.0$, HellaSwag reaches 31.4\% coverage and retains $+6.9$ points of
excess ($p=.001$); it has no significant excess at $\tau=0.5$ or $0.25$.
The first three benchmarks reach only 2.0--7.6\% coverage at $\tau=1.0$.
WinoGrande has no significant positive excess in any scenario.
Coverage must also be read alongside the actual DIF contrast: at $\tau=0.5$,
HellaSwag has 1,072 pairs in the first fold but 249 in the second; approximately
78.7\% of second-fold pairs have identical audit responses, with a median
final-refit DIF difference of zero.

\paragraph{Interpretation and reporting boundary.}
Optimizing over the full estimated profile preserves the original qualitative
pattern under an unconstrained paired-mixture reference, but does not hold
capability composition fixed. Stringent calipers reveal limited replacement
support under this particular distance, within-cell restriction, and
without-replacement design; they do not establish an intrinsic separation of
item populations or a unique family-dependence mechanism. Even close matches
would control only the estimated linear spectral profile, leaving unmeasured or
nonlinear capabilities, estimation error, and possible dependence of the final
spectral basis on the original purification procedure unresolved.
The independent diagnostic record retains all 25 benchmark--scenario results,
including nonsignificant settings; the tables above display the unconstrained
and $0.5$-logit scenarios, with the $1.0$-logit results summarized in text.

\end{document}